\documentclass{article}

\PassOptionsToPackage{numbers}{natbib}

\usepackage[preprint]{neurips_2022}

\usepackage[utf8]{inputenc} 
\usepackage[T1]{fontenc}    
\usepackage{hyperref}       
\usepackage{url}            
\usepackage{booktabs}       
\usepackage{amsfonts}       
\usepackage{nicefrac}       
\usepackage{microtype}      
\usepackage{multirow}
\usepackage{listings}
\usepackage[pdftex]{graphicx}
\usepackage[table,xcdraw,dvipsnames]{xcolor}
\usepackage{subcaption}
\usepackage{algorithm}
\usepackage{booktabs}
\usepackage{enumitem}
\usepackage{amsmath,amssymb}

\usepackage{pifont}
\newcommand{\cmark}{\ding{51}}%
\newcommand{\xmark}{\ding{55}}%

\usepackage{authblk}

\lstdefinestyle{examples}{
    frame=single,
    framerule=0pt,
    basicstyle=\ttfamily\footnotesize,
    backgroundcolor=\color{gray!10!white},
    columns=flexible,
    breaklines=false,
    keepspaces=false,
}

\lstdefinestyle{python}{
    frame=single,
    framerule=0pt,
    language=Python,
    basicstyle=\ttfamily\footnotesize,
    keywordstyle=\bfseries,
    backgroundcolor=\color{gray!10!white},
    commentstyle=\itshape\color{blue!60},
    columns=flexible,
    breaklines=true,
    keepspaces=true,
    showspaces=false
}

\lstdefinestyle{custom}{
    otherkeywords={<comments>,<python>,<eoc>,<description>,<descr>},
    frame=single,
    framerule=0pt,
    basicstyle=\ttfamily\footnotesize,
    keywordstyle=\bfseries\color{purple!65},
    backgroundcolor=\color{gray!10!white},
    keepspaces=true,
    breaklines=true,
    tabsize=4,
    columns=flexible,
    breakindent=0pt,
    numbers=none,
    tab=\t,
    showspaces=false, 
}

\newcommand{\codePLM}{\textsc{PanGu-Coder}}
\newcommand{\panggualpha}{\textsc{PanGu-$\alpha$}}

\title{\codePLM: Program Synthesis with Function-Level Language Modeling}

\author[1\thanks{Equal contributions}]{Fenia Christopoulou}
\author[1$^*$]{Gerasimos Lampouras}
\author[1$^*$]{Milan Gritta}
\author[1$^*$]{Guchun Zhang}
\author[1$^*$]{\authorcr Yinpeng Guo}
\author[2$^*$]{Zhongqi Li}
\author[2$^*$]{Qi Zhang}
\author[1]{Meng Xiao}
\author[2]{Bo Shen}
\author[2]{Lin Li}
\author[2]{\authorcr Hao Yu}
\author[2]{Li Yan}
\author[1]{Pingyi Zhou}
\author[1]{Xin Wang}
\author[2\thanks{Corresponding authors: \texttt{\{mayuchi1,ignacio.iacobacci,wangyasheng\}@huawei.com}}]{Yuchi Ma}
\author[1$^\dagger$]{Ignacio Iacobacci}
\author[1$^\dagger$]{\\Yasheng Wang}
\author[2]{Guangtai Liang}
\author[1]{Jiansheng Wei}
\author[1]{Xin Jiang}
\author[2]{\authorcr Qianxiang Wang} 
\author[1]{Qun Liu}
\affil[1]{Huawei Noah's Ark Lab}
\affil[2]{Huawei Cloud}

\begin{document}

\maketitle

\begin{abstract}


We present \codePLM, a pretrained decoder-only language model adopting the \panggualpha~architecture for text-to-code generation, i.e. the synthesis of programming language solutions given a natural language problem description. 
We train \codePLM~using a two-stage strategy: 
the first stage employs Causal Language Modelling (CLM) to pre-train on raw programming language data, while the second stage uses a combination of Causal Language Modelling and Masked Language Modelling (MLM) training objectives that focus on the downstream task of text-to-code generation and train on loosely curated pairs of natural language program definitions and code functions. Finally, we discuss \codePLM-FT, which is fine-tuned on a combination of competitive programming problems and code with continuous integration tests.
We evaluate \codePLM~with a focus on whether it generates functionally correct programs and demonstrate that it achieves equivalent or better performance than similarly sized models, such as CodeX \cite{codex}, while attending a smaller context window and training on less data.

\end{abstract}

\section{Introduction}

Increasingly more large pre-trained language models
~\cite{radford2018improving,devlin-etal-2019-bert,raffel-etal-2020-exploring,han-etal-2021-models} based on the transformer~\cite{vaswani2017attention} architecture have been proposed and shown to achieve state-of-the-art results on a variety of Natural Language Processing (NLP) tasks.
Lately, such models have been adapted to more specific language domains, e.g. to the Biomedical~\cite{huang-etal-2019-clinicalbert,lee2020biobert,DBLP:journals/corr/abs-1903-10676}, Legal~\cite{chalkidis-etal-2020-legal}, Cyber Security~\cite{https://doi.org/10.48550/arxiv.2204.02685}, and Finance~\cite{araci-2019-finbert} domains,
while simultaneously expanding to include signals from modalities other than natural language, e.g. Vision~\cite{DBLP:journals/corr/abs-1906-02940,chen2022when,pmlr-v119-chen20s,carion2020end,touvron-etal-2021-training,dosovitskiy2021an}, Proteins~\cite{brandes2022proteinbert}, Time Series~\cite{DBLP:journals/corr/abs-2001-08317, 10.1145/3394486.3403118,qin2022tsbert} and Code~\cite{kanade2020pretrained,wang-etal-2021-codet5,feng-etal-2020-codebert,guo-etal-2022-unixcoder,phan-etal-2021-cotext}.

In this work, we focus on pre-trained language models specifically created for text-to-code generation, i.e. the task of program synthesis from Natural Language (NL) descriptions (e.g. problem definitions or docstrings). While some of the proposed models for this task adapt the encoder-decoder architecture \cite{alpha_code}, most are trained as decoder-only transformer models \cite{codex, nijkamp2022conversational, fried2022incoder}. The encoder-decoder architecture requires a clear distinction and association between input and output. In contrast, a single component (encoder-only or decoder-only) architecture treats inputs as a continuous sequence, and as such is well-suited for training over vast amounts of raw data. 
Regardless of architecture, most recent work is further trained on large amounts of code data retrieved from GitHub\footnote{github.com}, Stack Exchange\footnote{stackoverflow.com} and other sources, with some being initialized from existing language models that were pre-trained on NL exclusively (e.g. initialised from GPT~\cite{brown2020language} or BERT~\cite{devlin-etal-2019-bert}). 

We present \codePLM, a pre-trained language model 
for text-to-code generation. \codePLM~follows the \panggualpha~architecture (see Figure~\ref{fig:pangu_alpha}) introduced by \citet{pangu_alpha}, which consists of a uni-directional
decoder-only transformer with an extra query layer stacked on top. At each time-step $t$, the query layer attends to the positional embedding $t + 1$ to determine the next token.
While \panggualpha~was proposed to handle both English and Chinese NL text, \codePLM~is currently focused on text-to-code generation from exclusively English prompts. \codePLM~only supports Python outputs at the moment, but the model is easily extendable to other languages.

We train \codePLM~using a two-stage strategy, 
with the first stage acting as pre-training on unsupervised raw programming language data. Natural language is included during this stage in the form of docstrings or inline comments, where available. To take full advantage of the raw data, we follow the training regime of existing decoder-only models and employ regular Causal Language Modeling (CLM)~\cite{radford2018improving} while treating all data as a continuous sequence.
The second stage of training is designed to focus on the downstream task of text-to-code generation and take advantage of the fact that it comprises distinct source and target sequences, i.e. the input NL problem definition and output code. 
As such, the second stage is focused exclusively on aligned pairs of NL and code. We experiment with various training objectives, including combinations of Causal and Masked Language Modelling (MLM) objectives, drawing inspiration from the training regimes of encoder-decoder architectures. 
Through this two-stage approach, our model is able to learn how general code structures relate to natural language through raw data in the first stage, and subsequently focus on how to best generate the correct output code given the NL input during the second stage. Finally, we also fine-tune \codePLM~using data that is more closely related to the target domain (Section \ref{finetuning}). We distinguish our stage-2 training from fine-tuning by the nature of the data used, i.e. extracted from possibly unaligned or noisy online sources vs. data more suited for text-to-code generation, e.g. retrieved from programming contests.



The remainder of this technical report is organized as follows. Section~\ref{sec:task_definition} defines the task of text-to-code generation formally.
In Section~\ref{sec:training_methodology}, we detail our training data, methodology and present zero-shot results and analysis. Similarly, Section~\ref{finetuning} presents our fine-tuning methodology, data and results. Finally, Section~\ref{sec:related_work} discusses related work.



\section{Task Definition}
\label{sec:task_definition}

\begin{table*}[h!]
\lstset{linewidth=9.5cm}
\centering
\begin{tabular}{>{\footnotesize\sc}ll}
\toprule
Problem Description &
\begin{lstlisting}[style=examples]
Return a greatest common divisor of two integers a and b \end{lstlisting} \\ \cmidrule{2-2}

Function Signature &
\begin{lstlisting}[style=examples]
def greatest_common_divisor(a: int, b: int) -> int:    \end{lstlisting} \\ \cmidrule{2-2} 

Unit Tests & 
\begin{lstlisting}[style=examples]
greatest_common_divisor(3, 5) = 1  
greatest_common_divisor(25, 15) = 5 \end{lstlisting} \\ \cmidrule{2-2}
 
Function Body &  
\begin{lstlisting}[style=examples]
while b:
    a, b = b, a % b 
return a \end{lstlisting} \\ 

\bottomrule
\end{tabular}
\caption{Examples taken from the HumanEval dataset.}
\label{tab:HE_example}
\end{table*}

The \codePLM~model we present in this report is specifically created for the task of text-to-code generation, i.e. to synthesize running code that successfully solves the problem described by an NL description. Table~\ref{tab:HE_example} shows an example from the HumanEval dataset~\cite{codex}, which is typical for the task; the input consists of an NL description of the problem, followed by the function signature of the solution (the function argument types and the expected return type may or may not be included). In addition, the problems may be accompanied by some unit tests, in the form of function calls with specific inputs and the corresponding expected outputs, e.g. \texttt{greatest\_common\_divisor(3, 5) = 1}. The model is then asked to produce the body of the code, and is evaluated against a number of held-out unit tests to ensure that the problem is properly solved. 


\section{Training Methodology}
\label{sec:training_methodology}





\begin{figure}[!t]
    \centering
    \includegraphics[width=0.85\linewidth]{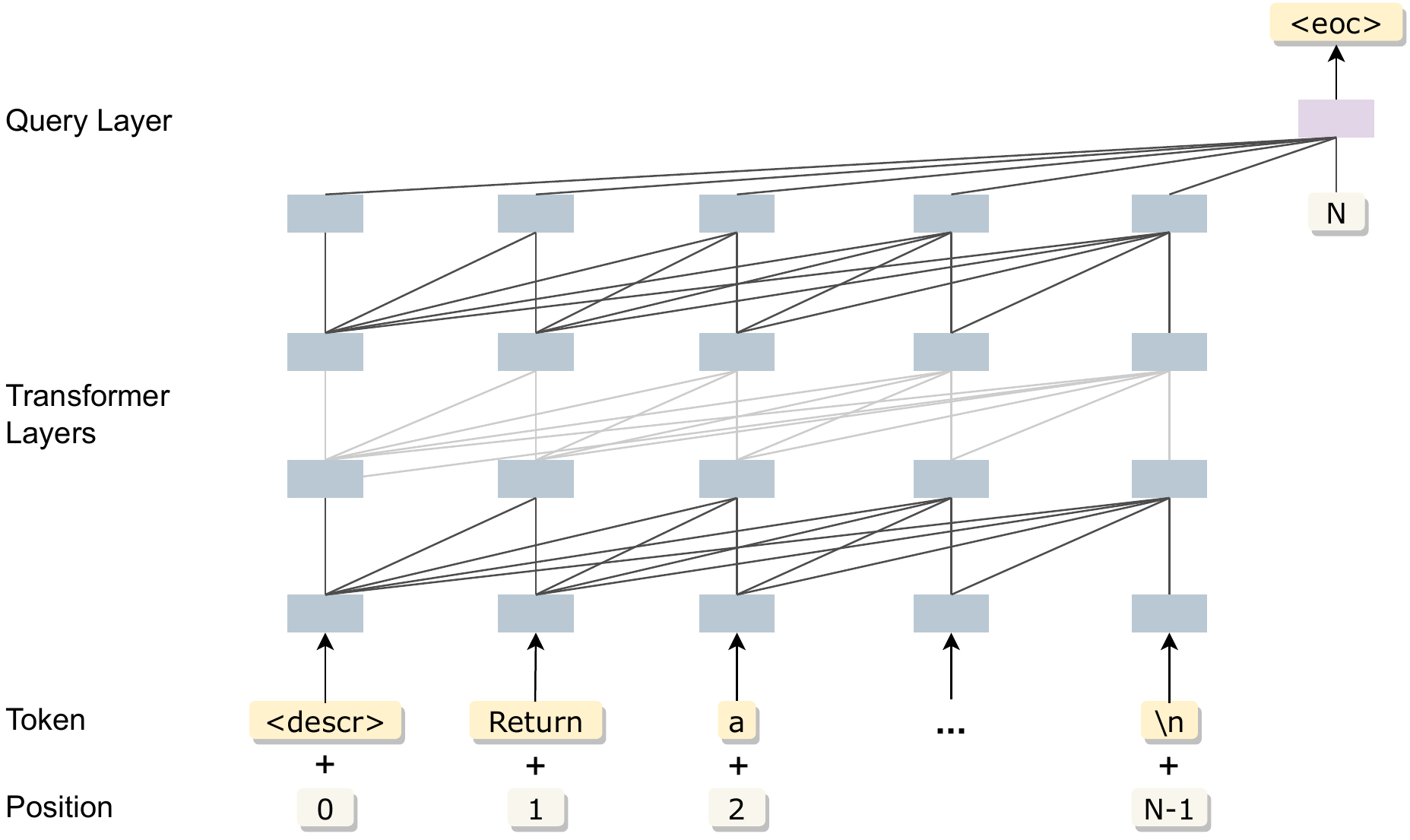}
    \caption{Schematic of the \panggualpha~architecture.}
    \label{fig:pangu_alpha}
\end{figure}

As mentioned in the introduction, \codePLM~uses \panggualpha~\cite{pangu_alpha} as its underlying architecture. \panggualpha~ was first developed to investigate the effect of large-scale pre-trained language models on Chinese NLP tasks. Its architecture (see Figure~\ref{fig:pangu_alpha}) was designed for scaling to hundreds of billions of parameters, and was implemented in MindSpore Auto-parallel\footnote{\url{https://www.mindspore.cn/en}} to allow for training parallelization across a cluster of 2,048 Ascend 910 AI processors\footnote{\url{https://e.huawei.com/en/products/servers/ascend}}.
The current version of \codePLM~model is implemented in Pytorch~\cite{paszke2019pytorch} and its training was done on Nvidia V100 GPU cards.

Similarly to GPT, \panggualpha~is a uni-directional autoregressive  decoder-only transformer with an additional attention layer on top, where an embedding $p_n \in \mathbb{R}^{d}$ indicating the next token position is used as the query vector in the attention mechanism. 

The attention weights in the extra layer are computed as follows:
\begin{equation}
    \alpha_h = p_n \; W_h^q \; W_h^{k^\top} \; H_L^\top,
    \label{eq:pangu_attention}
\end{equation}
where $W_h^q, W_h^k \in \mathbb{R}^{d \times d/N_h}$ are projection matrices, and $H_L \in \mathbb{R}^{V \times d}$ corresponds to the token representations obtained from the top transformer layer, with $h$ representing the index of the attention head, $d$ being the hidden dimension, $N_h$ the number of attention heads, and $V$ the vocabulary size.




\begin{table*}[!t]
	\small
	\centering
	\scalebox{0.9}{
	    \begin{tabular}{>{\sc}l>{\sc}c>{\sc}c>{\sc}r>{\sc}c>{\sc}c>{\sc}c}
		    \toprule
	        \multirow{2}{*}{Model} 
	        & \# Layers & Hidden size & \multicolumn{1}{c}{FFN size} & \# Heads & Context Size & Vocab \\
	        & ($L$) & ($d$) & \multicolumn{1}{c}{($d_\text{ff}$)} &  ($N_h$) & ($n_\textsc{cntx}$) & ($n_\textsc{vocab}$) \\
	        \midrule
            \codePLM~317 M & 24 & 1,024 &  4,096 & 16 & 1,024 & 41,865 \\
            \codePLM~2.6 B & 32 & 2,560 & 10,240 & 32 & 1,024 & 41,865 \\
			\bottomrule
		\end{tabular}
    }
	\caption{\codePLM~model sizes and configurations.} 
	\vspace{-5mm}
    \label{tab:model_sizes}
\end{table*}

For Chinese NLP tasks, multiple sized models were trained up to 200B parameters; for programming language modeling, we test the following model configurations, as shown in Table \ref{tab:model_sizes}. 
In the remainder of this section, we 
discuss the data used for training and evaluation, and show how we use the \panggualpha~model, together with its accompanying tokenizer and vocabulary, to train it on code-specific data using various strategies. The section concludes with an analysis of the effect of different decoding strategies on the models' zero-shot performance.

\subsection{Data}
\label{sec:data}

\paragraph{Collection}
The initial dataset was collected through GHTorrent \cite{Gousi13},
an online tool that 
collects and stores Github public event metadata.
In an effort to make our pre-training data comparable to previous work, we focused exclusively 
on Github repositories established before May 2021. 
The download process occurred in between March 2022 and May 2022 using a 280 nodes Spark cluster.

For model pre-training, about 65 million Python files were extracted, totaling a size of 380 GB. 
All files were processed on a Spark cluster, and a rowKey of MD5 (generated from each file's content) was used as the unique index of the NoSQL database to remove duplicate files. 
After this we ended up with about 40 million files, and the total size was reduced to 185 GB. To establish data quality, we only kept the collected files that met the following criteria:
\begin{itemize}[leftmargin=0.5cm,nolistsep]
\item [a)] the file size is under 1MB; 
\item [b)] the code is Python3 compatible, as determined through its Abstract Syntactic Tree (AST); 
\item [c)] there are fewer than 100 characters per line on average; 
\item [d)] and there are fewer than 1,000 characters in any single line. 
\end{itemize}

This process further reduced the number of data
to about 147 GB in size. 

\begin{figure}[t!]
    \centering
    \includegraphics[width=\linewidth]{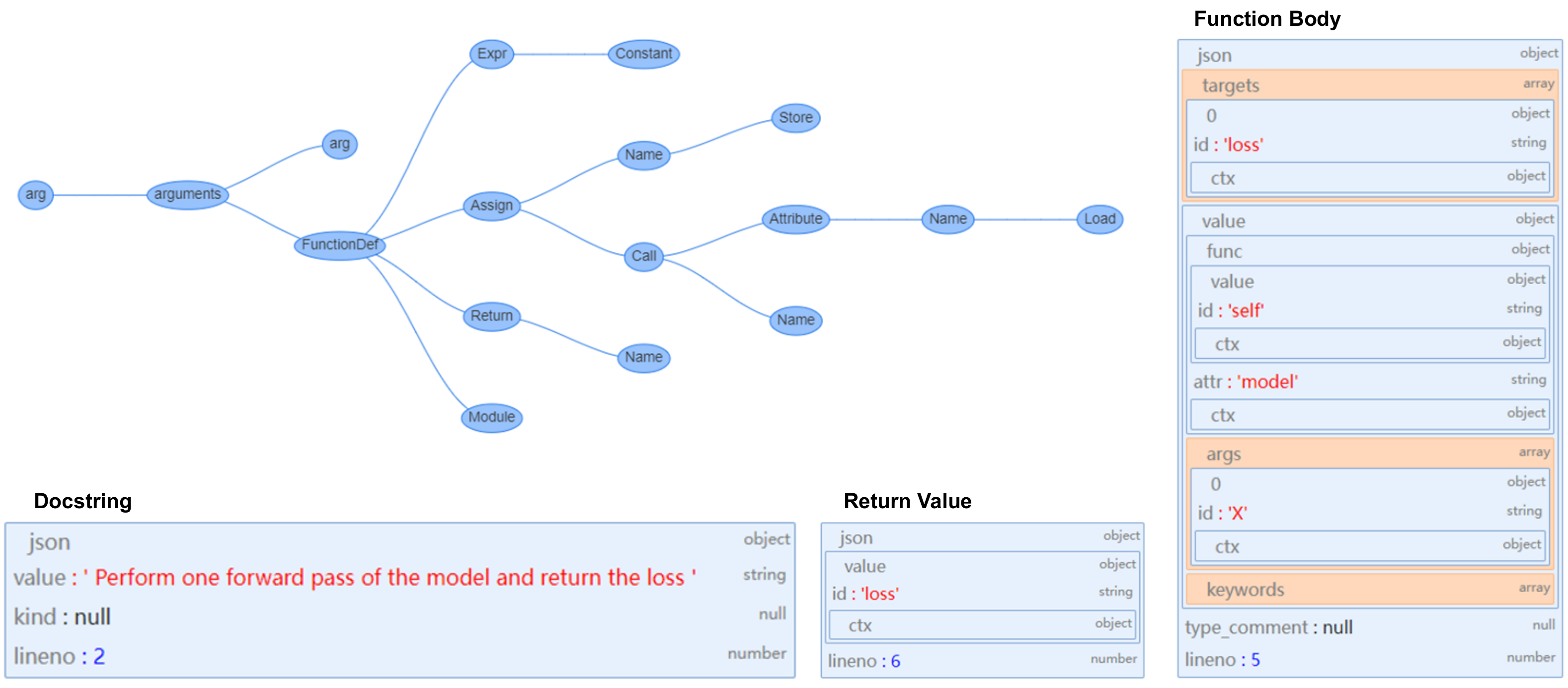}
    \caption{Example of the Abstract Syntax Tree of an extracted function.}
    \label{fig:ast_parse}
    \vspace{-5mm}
\end{figure}

\paragraph{Pre-processing} To format the data for text-to-code generation, as described in Section~\ref{sec:task_definition}, we need to extract the problem descriptions, the corresponding function signatures and bodies. We apply AST parsing on each of the remaining files and extract functions and corresponding docstrings\footnote{While we also extract and train on classes, we will only refer to functions as the process is identical.} (see  Figure~\ref{fig:ast_parse}). A docstring is a snippet of natural language text that immediately follows the definition of a function or class, 
usually enclosed with the multi-line annotation in Python (\texttt{"""}), as shown below. Exploiting the AST's structure, the docstring
node is located and the text is extracted.
We then collapse the AST structure, ignoring the docstring node, to re-obtain the function signature and body. 
The benefit of this process is that the code is well formatted with all meaningless empty lines and spaces removed. 
Finally, we apply the de-duplication process to the extracted function bodies as well.

For training, we use these docstrings in lieu of NL problem descriptions, as we consider them to be the closest readily available alternative. 

\begin{lstlisting}[style=python]
def _forward(self, X):
    """ 
    Perform one forward pass of the model and return the loss 
    """
    loss = self.model(X)
    return loss
\end{lstlisting}

We create our training samples using two data formats, depending on whether a corresponding docstring is available for the relevant function body. 
When a docstring exists, our training samples are constructed as
\texttt{<descr> docstring <python> code <eoc>}: 
\begin{lstlisting}[style=custom]
<descr> Perform one forward pass of the model and return the loss <python>\n def _forward(self, X):\n   loss = self.model(X)\n    return loss <eoc>
\end{lstlisting}

When there is no docstring available, we omit the \texttt{<descr>} segment and assemble the samples as \texttt{<python> code <eoc>}:
\begin{lstlisting}[style=custom]
<python>\n def _forward(self, X):\n    loss = self.model(X)\n    return loss <eoc>
\end{lstlisting}
Here, the \texttt{<descr>} symbol marks the beginning of the problem description, \texttt{<python>} indicates the beginning of the code solution, while the \texttt{<eoc>} symbol (end-of-code) marks the end of the docstring-code or code-only instance.
During the first stage of training, we use both data formats (code-only and docstring-code), while only the docstring-code format is used for the second stage.

\label{sec:eval_data}

\paragraph{Evaluation} 
In order to evaluate our models, we make use of two of the most commonly used datasets for evaluating the functional correctness of generated programs, namely HumanEval~\cite{codex} and the Mostly Basic Programming Problems~\cite[MBPP]{austin2021program} datasets. 
HumanEval\footnote{\url{https://github.com/openai/human-eval}} contains 164 handcrafted Python problems accompanied by a set of held-out unit tests (average of 7.7 unit tests per problem), \textit{all of which} must pass in order to count as a successful solution. We note that HumanEval may also provide some unit tests as a part of the problem description, but they are distinct from the unit tests used for evaluation. HumanEval is a good estimator of zero-shot model performance as its handcrafted problem descriptions ensure little to no overlap with any pre-training data that are automatically extracted from online resources. 
In a similar vein,  MBPP is comprised of 974 programming problems (474 train, 500 test) designed to be solved by entry-level Python programmers.



\paragraph{Evaluation Metrics}
In order to estimate model performance, we sample $n$ programs/solutions per problem and calculate pass@$k$ for $k$ = [$1$, $10$, $100$]. To avoid issues of high variance, we use the unbiased estimator of pass@$k$ introduced by \citet{codex}. In all the experiments we use a sample size $n=200$, unless otherwise stated.

An issue we observed in preliminary evaluation arises from the fact that the pre-training and fine-tuning data are focused exclusively on function signatures, bodies and their preceding docstring. As such, the code generated by our models often misses the required dependencies. 
To accurately estimate performance, whenever there is a missing dependency, the solution is re-evaluated with the missing library automatically imported before the function call, essentially simulating the auto-import function of many integrated development environments (i.e. IDEs).

\subsection{Training Stages and Objectives}
\label{sec:pretrain_objectives}
We train \codePLM~using a 2-stage process, 
as detailed below. 
Formally, we denote the docstring and code output as a single sequence of tokens $X = X_D + X_C = \{d_1, ..., d_{N_D}, c_1, ..., c_{N_C}\}$, with $X_D$ and $X_C$ the tokens corresponding to the \textit{docstring} and \textit{code}, respectively.

\subsubsection{Stage-1 Training}
\label{sec:stage_1_training}
During the first stage of training, the \codePLM~model is initialised with random weights and trained on Python code via Causal Language Modeling (CLM), also known as next token prediction, using the formats introduced in Section \ref{sec:data}. 
\begin{equation}
    \mathcal{L}_\textsc{clm}(X) = \mathcal{L}_\textsc{clm}(X_D, X_C) = - \frac{1}{N} \sum_{n=1}^{N} \log p(x_n| x_{i \leq n-1}; \theta),
    \label{eq:clm}
\end{equation}
where $N = N_D + N_C$ indicates the total number of tokens in the input sequence $X$ (with $N_D$, $N_C$ being the number of \textit{docstring} and \textit{code} tokens, respectively)
and $\theta$ corresponds to the model parameters. In the following equations $\theta$ is omitted for brevity.

\begin{figure}[t!]
    \centering
    \begin{subfigure}[b]{0.53\linewidth}
        \includegraphics[width=\linewidth]{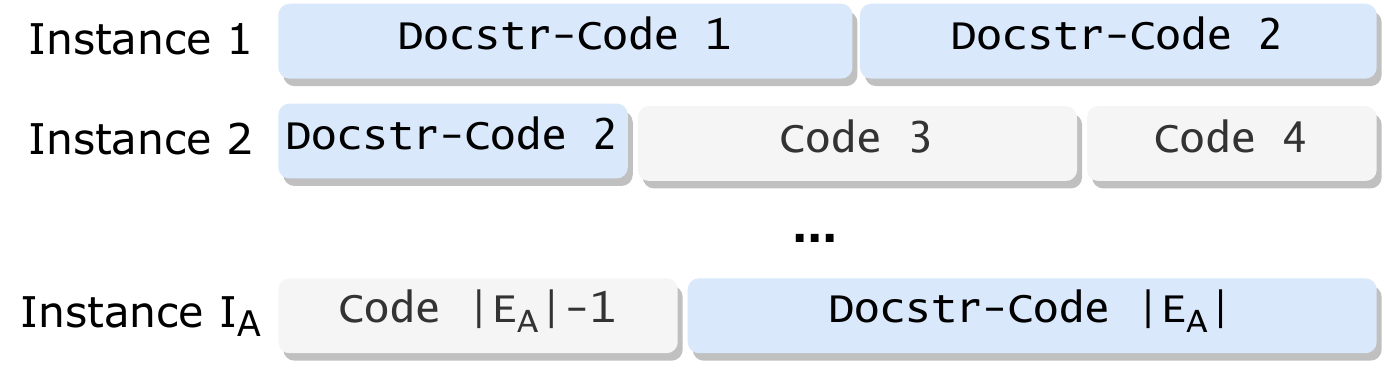}
        \caption{Stage-1 instance formation, where all available data are concatenated and split at a given length.}
        \label{fig:stage1_concat}
    \end{subfigure}\hfill%
    \begin{subfigure}[b]{0.45\linewidth}
        \includegraphics[width=\linewidth]{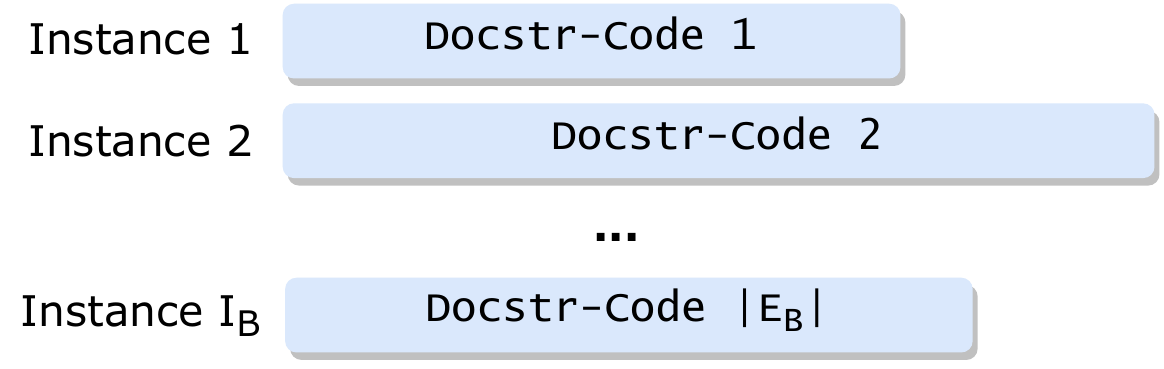}
        \caption{Stage-2 instance formation, with each docstring-code pair fed separately to the model.}
        \label{fig:stage2_pair}
    \end{subfigure}
    \caption{Input formats during stage-1 and stage-2 training.}
\end{figure}

In order to formulate the model input, we concatenate all $E_A$ training examples as a single sequence (both code-only and docstring-code examples) and generate training instances by splitting the concatenated sequence into $I_A \leq |E_A|$ chunks of $1,024$ subwords, including the inserted special tokens, as shown in Figure \ref{fig:stage1_concat}. 
The model is then trained for 188 billion tokens in total.


\subsubsection{Stage-2 Training}
\label{sec:stage_2_training}
For the second stage, we form the model inputs by exclusively considering $|E_B| \leq |E_A|$ docstring-code examples and treating each as a single training instance $I_B = |E_B|$, as shown in Figure \ref{fig:stage2_pair}. To further reduce noise in the data, we removed edge cases where the docstring was shorter than 19 words, the function body longer than 400, or where their length ratio was greater than 32; these values were empirically determined through observation of the curated datasets' statistics.

\begin{figure}[h!]
    \centering
    \includegraphics[width=\linewidth]{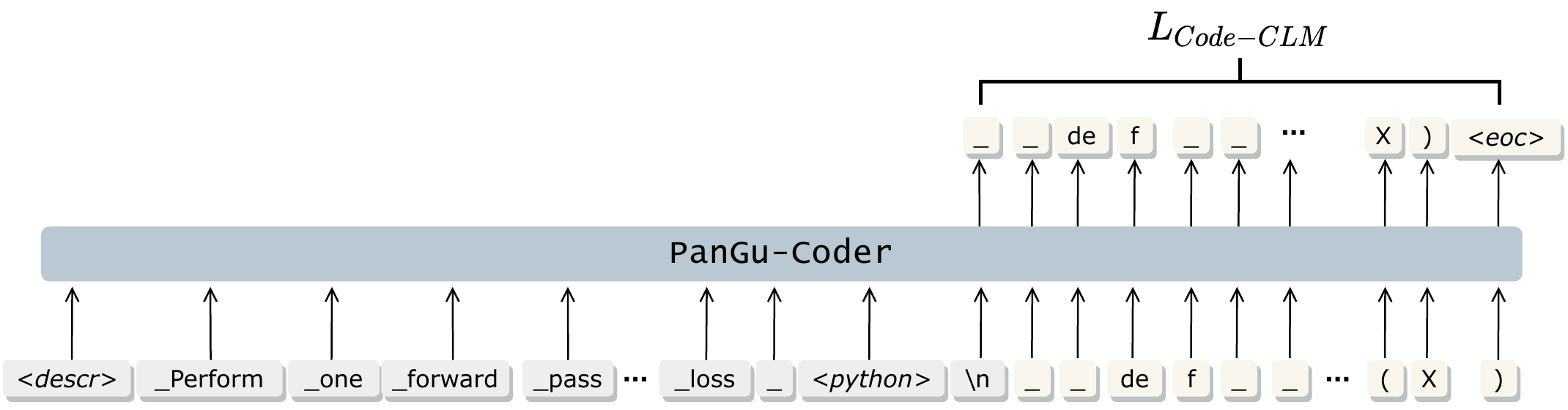}
    \caption{\textsc{Code-CLM}: Causal Language Modeling over code-only tokens.}
    \label{fig:clm_code}
\end{figure}

We explore several objectives for stage-2 model training on Python code, 
each consisting of mixtures of joint losses, focusing independently on the docstring and code subsequences. The combinations are primarily motivated by the shift in focus to the downstream task of text-to-code generation during stage-2. We present the individual losses below:

\paragraph{Code-CLM: Causal Language Modeling on Code} 
This loss is computed by applying CLM exclusively on the code subsequence, hence it is named \textsc{Code-CLM}. 
Enforcing this objective during pre-training brings us closer to the target objective of the downstream task. 
\begin{equation}
    \mathcal{L}_\textsc{Code-CLM}(X) = - \frac{1}{N_C} \sum_{n=1}^{N_C} \log p(c_n|c_{i \leq n-1}, d_1, ..., d_{N_D}),
    \label{eq:code_clm}
\end{equation}
As shown in Equation (\ref{eq:code_clm}), and depicted in Figure \ref{fig:clm_code}, each code token $c_i$ is predicted based on all previous tokens, including the tokens of docstring $d \in X_D$. 

\begin{figure}[h!]
    \centering
    \begin{subfigure}[t]{0.49\linewidth}
        \includegraphics[width=\linewidth]{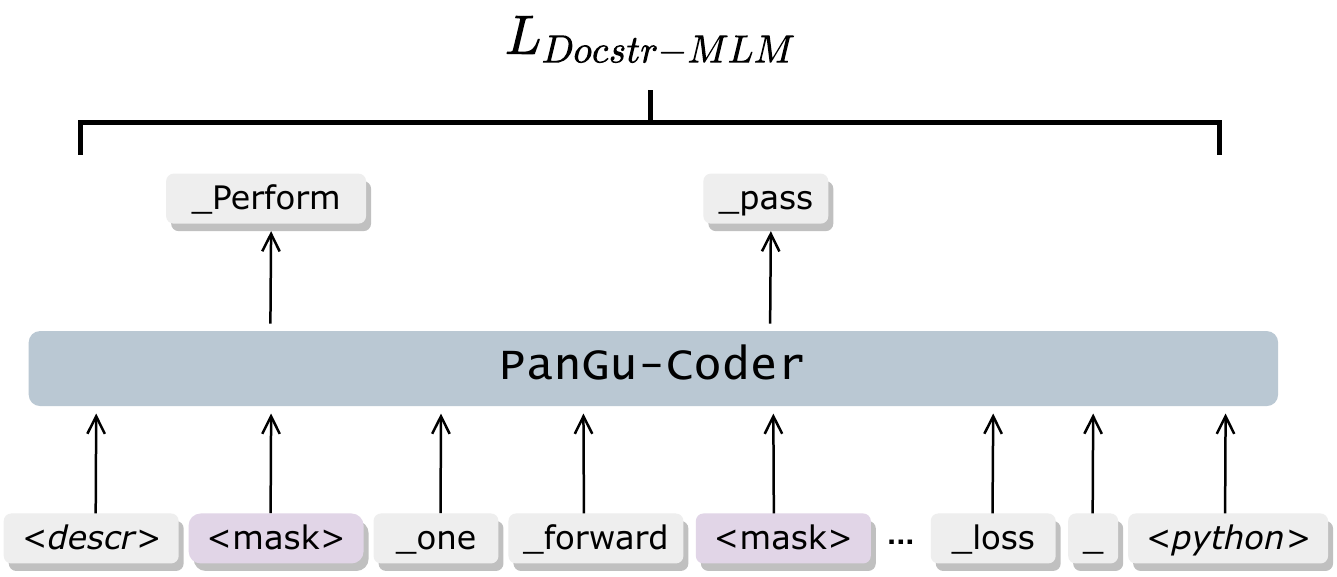}
        \caption{\textsc{Docstr-MLM}: Masked Language Modeling on docstring.}
        \label{fig:docstr_mlm}
    \end{subfigure}\hfill%
    \begin{subfigure}[t]{0.49\linewidth}
        \includegraphics[width=\linewidth]{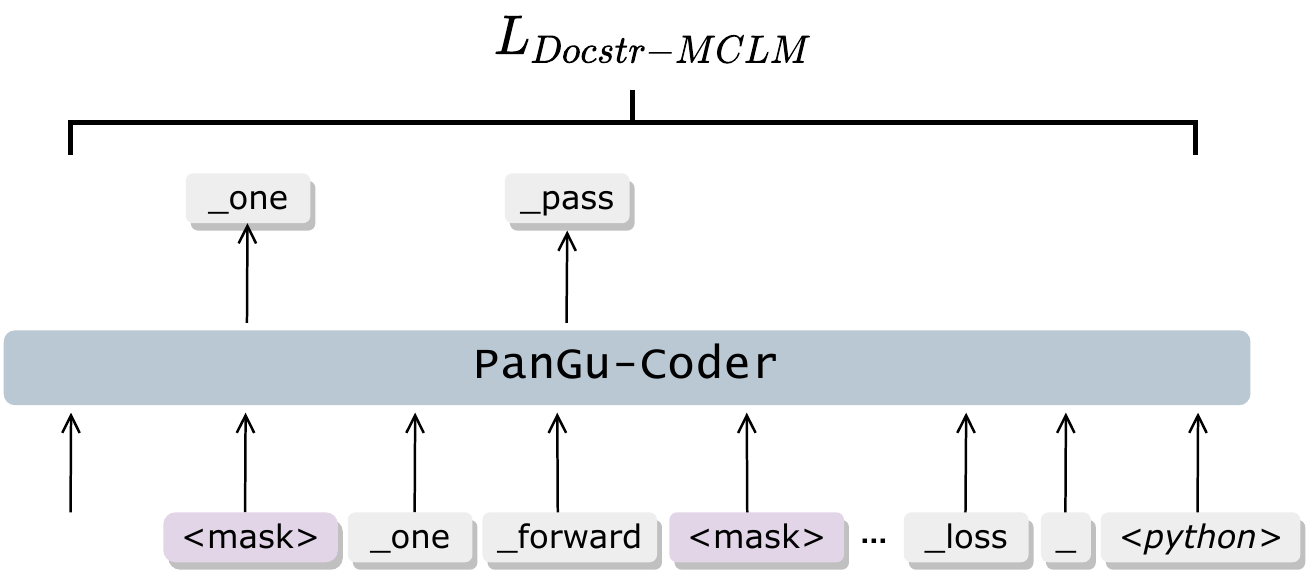}
        \caption{\textsc{Docstr-MCLM}: Masked Causal Language modeling on docstring.}
        \label{fig:docstr_masked_clm}
    \end{subfigure}
    \caption{Masked Language Modeling modifications applied on the docstring.}
\end{figure}

\paragraph{Docstr-MLM: Masked Language Modeling on Docstring}
Since the down-stream task is not reliant on next word prediction for the docstring, this loss calculates standard Masked Language Modeling (MLM) exclusively on the docstring (\textsc{Docstr-MLM}) as depicted in Figure \ref{fig:docstr_mlm}. 
Specifically, a few random $M < N_D$ tokens in the docstring are replaced with a mask, a random token or the same token with $0.8$/$0.1$/$0.1$ chance, respectively, similar to \citet{devlin-etal-2019-bert}.

In contrast to models that perform MLM in a bidirectional fashion, using a single decoder network (namely prefix LMs)~\citep{dong-etal-2019-unilm,bao-etal-2020-unilmv2,guo-etal-2022-unixcoder}, in our case we do not change the underlying attention mechanism of the model. As a result, the masked docstring tokens are predicted by only attending to previous ones.

Finally, we need to underline that in this objective, we shift from next token prediction to same (masked) token prediction solely on the docstring.
\begin{equation}
    \mathcal{L}_\textsc{Docstr-MLM}(X_D) = - \frac{1}{M} \sum_{m=1, m \in N_D}^{M} \log p(d_m|d_{i \leq m})
    \label{eq:docstr_mlm}
\end{equation}

\paragraph{Docstr-MCLM: Masked Causal Language Modeling on Docstring}
Similarly to the previous one, in this loss we randomly mask tokens in the docstring, but instead of predicting the masked tokens, we employ causal language modeling on the masked docstring.
To make the task easier and the training more efficient, we calculate the docstring loss over a few randomly selected tokens $J < N_D$, with probability $0.15$.
We name this loss \textsc{Docstr-MCLM} as the input docstring includes \textit{masked} tokens while computing a causal language modeling loss on it.
An example of the loss is shown in Figure \ref{fig:docstr_masked_clm}. 
\begin{equation}
\label{eq:docstr_mclm}
    \mathcal{L}_\textsc{Docstr-MCLM}(X_D) = - \frac{1}{J} \sum_{j=1, j \in N_D}^{J} \log p(d_j|d_{i \leq j-1})
\end{equation}
where $J$ corresponds to a number of random tokens in the docstring, selected for prediction and $d_i$ is the $i$-th docstring token.

\paragraph{Training objectives} In Table~\ref{tab:losses}, we summarise the objectives and the corresponding joint losses with which \codePLM~is trained.
The first objective uses exclusively the \textsc{Code-CLM} loss and no loss is calculated over the docstring. As \codePLM~is meant to generate only code, since the docstring of the problem will be provided as input, it is preferable to prioritise the next token prediction loss of the code subsequence over the docstring.
In the next two objectives, we re-introduce different computations of the loss on the docstring part of the sequence. Typically, in prefix LMs both input segments (docstring and code) are masked but only tokens in the second segment (code) are predicted. Here, instead, we apply masking and prediction on the first segment, while on the second we predict tokens one-by-one without any masking.

\begin{table}[h!]
    \centering
    \begin{tabular}{ll}
    \toprule
    \textsc{Objective} & \textsc{Loss} \\ \midrule
      Code-CLM  
      & $\mathcal{L} = \mathcal{L}_\textsc{Code-CLM}(X)$ \\
      
      Docstr-MCLM + Code-CLM 
      & $\mathcal{L} = \mathcal{L}_\textsc{Docstr-MCLM}(X_D) + \mathcal{L}_\textsc{Code-CLM}(X)$ \\
      
      Docstr-MLM + Code-CLM 
      & $\mathcal{L} = \mathcal{L}_\textsc{Docstr-MLM}(X_D) + \mathcal{L}_\textsc{Code-CLM}(X)$ \\
    \bottomrule \\
    \end{tabular}
    \caption{Pre-training losses for different objectives.}
    \label{tab:losses}
\end{table}
\vspace*{-0.5cm}

\subsection{Tokenization}
\label{sec:tokenization}


We inherit the tokenization model and corresponding vocabulary from \panggualpha, which was built using SentencePiece~\cite{kudo-richardson-2018-sentencepiece} and contains 70K subtokens. However, the original \panggualpha~vocabulary supports both Latin and Chinese characters; since there is no need for the latter in the applied datasets, our model's vocabulary consists only of 42K subtokens. After the input sequences are constructed using the data formats described in Section \ref{sec:data}, we apply the tokenization model to split each sequence into subtokens as the sole pre-processing step before training or testing. 

When the same tokenization strategy is used for the whole sequence, some embeddings may be shared between the docstring and code subsequences. In Stage-1 training, the embeddings are shared across docstring and code, but in Stage-2 training we experimented with both sharing and separating the embeddings used in each subsequence.


\subsection{Pre-training Results and Analysis}
\label{sec:pretraining-results-analysis}

\paragraph{Training details}
We train models with a batch size of $256$, using the Adam optimizer~\citep{adam} with $\beta_1 = 0.9$, $\beta_2 = 0.95$ and weight decay of $0.01$. We employ a cosine decay scheduler with maximum and minimum learning rates 1e-5 and 5e-6, respectively. The gradients are clipped at $3.0$ during stage-1 and $1.0$ during stage-2.
For stage-1 training, the model is initialised with random weights and trained from scratch with a warmup ratio of 1\% for 188 billion tokens in total. For stage-2, we initialise the model from a checkpoint obtained during stage-1, reset the optimizer, keep the same hyper-parameters and continue training for a maximum of 1M steps, corresponding to 42 billion tokens.

\paragraph{Generation}
As our models are based on the \panggualpha~architecture, we follow the standard decoding / generation process used for auto-regressive language models. Generation adopts a prompt that is similar to the data format used during the pre-training stages (c.f. Section~\ref{sec:data}), with only the function signature provided and not the body, i.e. \texttt{<descr> docstring <python> signature}. Due to the nature of our applied tasks, we assume that there is always a docstring available during generation. We remove any superfluous whitespaces and line breaks from the problem descriptions.

\begin{lstlisting}[style=custom]
<descr> Perform one forward pass of the model and return the loss <python>\n def _forward(self, X):\n
\end{lstlisting}

Formally, the prompt is a sequence of tokens $P = P_D + P_S = \{d_1, ..., d_{N_D}, s_1, ..., s_{N_S}\}$ where $|N_D|, |N_S|$ are the number of tokens in the docstring and signature subsequences. The model then generates a continuation $C'$ of the prompt in a left-to-right manner, decoding one token at a time while attending on the previous tokens in the sequence. The generation continues until either a special end token (i.e. the \texttt{<eoc>} token) or a maximum sequence length is reached. 
\begin{equation}
C'(x_t) = \text{\codePLM~}(d_{1}, ..., d_{N_D}, s_1, ..., s_{N_S}, c'_{0}, ..., c'_{t-1}).
\end{equation}
As is standard in such models, the decoding can be performed through a number of stochastic strategies. Here we use temperature scaling ($t$) and nucleus sampling ($p$) \cite{HoltzmanBDFC20}, and 
following previous work, we report results from the best settings in each configuration. We analyse the effect of different decoding strategies and prompt variations in the later subsections.

\subsubsection{Zero-shot Results}
Tables \ref{tab:human_eval_results} and \ref{tab:mbpp_main_results} summarise our best results on the HumanEval and MBPP datasets respectively.
For the 317M parameter model, we use 
$p=0.4$, $t=0.2$ for pass@$1$, and $p=0.8$, $t=0.95$ for pass@$10$ and pass@$100$. 
For the 2.6B model, we use $p=0.4$, $t=0.2$ for pass@$1$, and $p=0.8$, $t=0.65$ for  pass@$10$ and pass@$100$. 

\definecolor{Gray}{gray}{0.9}
\begin{table}[t!]
    \centering
    \scalebox{0.87}{
    \begin{tabular}{p{3.2cm}rcrrr|rrr}
        \toprule
        \multirow{2}{*}{\textsc{Model}} 
        & \multirow{2}{*}{\textsc{size}} 
        & \multirow{2}{*}{$n_\textsc{cntx}$}
        & \multirow{2}{*}{$n_\textsc{vocab}$}
        & \multicolumn{1}{c}{\textsc{data}}
        & \multicolumn{1}{c|}{\textsc{train}}
        & \multicolumn{3}{c}{\textbf{\textsc{HumanEval}} (\%)}  \\
        &  &  & & \multicolumn{1}{c}{\textsc{(gb)}} & \multicolumn{1}{c|}{\textsc{tokens}}
        & \small \textsc{pass@$1$} & \small\textsc{pass@$10$} & \small\textsc{pass@$100$} \\ \cmidrule{1-1} \cmidrule(lr){2-6} \cmidrule(lr){7-9}
        
        \multirow{1}{*}{\textsc{GPT-Neo} \cite{gpt-neo}}  
        & 125 M & 2,048 & 50 K & 825 & 300 B & 0.75 & 1.88 & 2.97 \\ \midrule
        
        \multirow{1}{*}{\textsc{Codex} \cite{codex}}
        & 300 M  & 4,096 & 50 K & 729 & 400 B & 13.17 & 20.37 & \textbf{36.27} \\ 
        \multirow{1}{*}{\textsc{AlphaCode} \cite{alpha_code}}
        & 302 M & 2,304  & 8 K &  715 & - & 11.60 & 18.80  & 31.80 \\
        \multirow{1}{*}{\textsc{CodeGen Multi} \cite{nijkamp2022conversational}}
        & 350 M &  2,048 & 50 K & 1,595  & 250 B  &  6.67 & 10.61 & 16.84 \\
        \multirow{1}{*}{\textsc{CodeGen Mono} \cite{nijkamp2022conversational}}
        & 350 M &  2,048 & 50 K & 1,812 & 325 B  & 12.76 & 23.11 & 35.19 \\
        \rowcolor{Gray}
        \multirow{1}{*}{\textbf{\textsc{\codePLM}}}
        & 317 M & 1,024 & 42 K & 147 & 211 B & \textbf{17.07} & \textbf{24.05} & 34.55 \\ \midrule
        
        \multirow{1}{*}{\textsc{Codex}}
        & 679 M  & 4,096 & 50 K & 729 & 400 B & 16.22 & 25.70 & 40.95 \\ 
        \multirow{1}{*}{\textsc{AlphaCode}}
        & 685 M & 2,304  & 8 K &  715 &  -    & 14.20 & 24.40  & 38.80 \\ \midrule
        
        \multirow{1}{*}{\textsc{AlphaCode}}
        & 1.1 B & 2,304  & 8 K &  715 & - & 17.10 & 28.20  & 45.30 \\
        \multirow{1}{*}{\textsc{GPT-Neo}}  
        & 1.3 B & 2,048 & 50 K & 825 & 380 B & 4.79 & 7.47 & 16.30 \\ \midrule
         
        \multirow{1}{*}{\textsc{Codex}}
        & 2.5 B  & 4,096 & 50 K & 729 & 400 B & 21.36 & 35.42 & \textbf{59.50} \\ 
        \rowcolor{Gray}
        \multirow{1}{*}{\textbf{\textsc{\codePLM}}}
        & 2.6 B & 1,024 & 42 K & 147 & 387 B & \textbf{23.78} & 35.36 & 51.24 \\
        \multirow{1}{*}{\textsc{CodeGen Multi}}
        & 2.7 B &  2,048 & 50 K & 1,595  & 500 B & 14.51 & 24.67 & 38.56 \\
        \multirow{1}{*}{\textsc{CodeGen Mono}}
        & 2.7 B &  2,048 & 50 K & 1,812 & 650 B & 23.70 & \textbf{36.64} & 57.01 \\
        \multirow{1}{*}{\textsc{GPT-Neo}}  
        & 2.7 B & 2,048 & 50 K & 825 & 420 B & 6.41 & 11.27 & 21.37 \\ 
        \midrule
        
        \textsc{GPT-J} \cite{mesh-transformer-jax} 
        & 6 B & 2,048 & 50 K & 825 & 402 B &  11.62 & 15.74 & 27.74 \\
        \multirow{1}{*}{\textsc{CodeGen Multi}}
        & 6.1 B &  2,048 & 50 K & 1,595  & 1 T & 18.20 & 28.70 & 44.90  \\
        \multirow{1}{*}{\textsc{CodeGen Mono}}
        & 6.1 B &  2,048 & 50 K & 1,812 & 1.3 T & \textbf{26.13} & \textbf{42.29} & \textbf{65.82} \\
        \textsc{InCoder} \cite{fried2022incoder} 
        & 6.7 B & 2,048 & 27.6 K & 216 & 52 B & 15.20 & 27.80 & 47.00 \\
        
        \bottomrule \\
    \end{tabular}
    }
    \caption{Pass@$k$ rates on the HumanEval dataset, among various models. Sizes are reported in thousands (K), millions (M), billions (B) and trillions (T).\protect\footnotemark}
    \label{tab:human_eval_results}
\end{table}

\begin{table}[t!]
    \centering
    \scalebox{0.88}{
    \begin{tabular}{lrcrrr|rrr}
        \toprule
        \multirow{2}{*}{\textsc{Model}} 
        & \multirow{2}{*}{\textsc{size}} 
        & \multirow{2}{*}{$n_\textsc{cntx}$}
        & \multirow{2}{*}{$n_\textsc{vocab}$}
        & \multicolumn{1}{c}{\textsc{data}}
        & \multicolumn{1}{c|}{\textsc{train}}
        & \multicolumn{3}{c}{\textbf{\textsc{MBPP}} (\%)}  \\
        &  &  & & \multicolumn{1}{c}{\textsc{(gb)}} & \multicolumn{1}{c|}{\textsc{tokens}}
        & \textsc{pass@$1$} & \textsc{pass@$10$} & \textsc{pass@$100$} \\ \midrule
        
        \textsc{InCoder} \cite{fried2022incoder} 
        & 6.7 B & 2,048 & 22.6 K & 216 & 52 B & 19.40 & - & - \\
        \midrule
        
        \rowcolor{Gray}
        & 317 M & 1,024 & 42 K & 147 & 211 B & 16.20 & 34.39 & 53.74 \\
        \rowcolor{Gray}
        \multirow{-2}{*}{\textbf{\textsc{\codePLM}}}
        & 2.6 B & 1,024 & 42 K & 147 & 387 B & 23.00 & 43.60 & 59.64 \\
        \bottomrule \\
    \end{tabular}
    }
    \caption{Pass@$k$ rates on the MBPP dataset.}
    \label{tab:mbpp_main_results}
\end{table}

Table \ref{tab:human_eval_results} reports a comparison of existing models, as well as our proposed \codePLM~on the HumanEval dataset, along with accompanied model and data sizes, the contextual window allowed ($c_\textsc{cntx}$), vocabulary size ($n_\textsc{vocab}$) and number of tokens the models were trained for.

For \textsc{Codex}~\citep{codex}, the total data size and number of trained tokens is calculated by considering the initial training of GPT-3~\cite{brown2020language} for 300 billion tokens on a collection of data equivalent to 570GB of compressed plain text. 
Similarly, for \textsc{CodeGen}~\citep{nijkamp2022conversational}, the dataset size was computed by accumulating The Pile~\cite{thepile} and BigQuery\footnote{\url{https://cloud.google.com/bigquery/public-data/}} for \textsc{CodeGen-Multi} and combining The Pile, BigQuery and BigPython~\citep{nijkamp2022conversational} for \textsc{CodeGen-Mono}, as models were trained sequentially on the three datasets. In order to calculate the number of training tokens for the \textsc{CodeGen} models we assume the batch size reported in the paper corresponds to tokens instead of instances.
Information about \textsc{GPT-Neo} and \textsc{GPT-J} was obtained via the model cards available\footnote{\url{https://huggingface.co/EleutherAI}}.
For \textsc{InCoder}, the vocabulary size was calculated as 55\% of GPT-2 vocabulary, based on \citet{fried2022incoder}.
For the rest of the models, explicit information was provided in the corresponding papers.
For all models, pass@$k$ rates are computed with $200$ samples, except for \textsc{AlphaCode} where the reported rates used $1,000$ samples.\footnote{We include the decoder-only baseline presented by \textsc{AlphaCode}, and not the encoder-decoder model, as HumanEval results are only reported on the former. The number of train tokens of this baseline are not reported.}

\codePLM~results in the best performance in the 300M family of models for pass@$1$ and pass@$10$.
For pass@$100$, {\codePLM} performs lower than \textsc{CodeGen-Mono} and \textsc{Codex},
but the latter has been trained on a 2x and 4x larger input context respectively, and for at least four times more data and more tokens.
Looking at the 2.6B models family, \codePLM~again achieves the best pass@$1$ performance. On the other hand, \codePLM~underperforms compared to \textsc{CodeGen-Mono} and \textsc{Codex} on pass@$10$ and pass@$100$, but similarly to the 300M family of models, these two have been trained with a larger context, on more data, and for more tokens.

\footnotetext{We did not include even larger scale models, since they would not be directly comparable with this work.}

Regarding the MBPP dataset, most other models do not report zero-shot results on it, with only \textsc{InCoder} reporting pass@$1$. \codePLM~2.6B outperforms \textsc{InCoder} even though it is less than half its size (2.6B vs 6.7B parameters). Furthermore, even though we are not able to make an apples-to-apples comparison with the \codePLM~317M model due to their size difference, it is interesting to note that it is only 3.2 points below \textsc{InCoder} on pass@$1$.

Overall we observe, and this is further confirmed through analysis in later sections, that \codePLM~performs better for lower $k$ values. We speculate that the gap in performance for pass@$100$ is a result of the small context size {\codePLM} has been trained on, that prevents the model from learning to generate long solutions, or solutions that have to attend over quite long descriptions. 





\subsubsection{Impact of Different Decoding Strategies}
\label{sec:hyper_effect}
\begin{figure}[h!]
    \centering
    \begin{subfigure}[b]{0.5\linewidth}
        \centering
        \includegraphics[width=\linewidth]{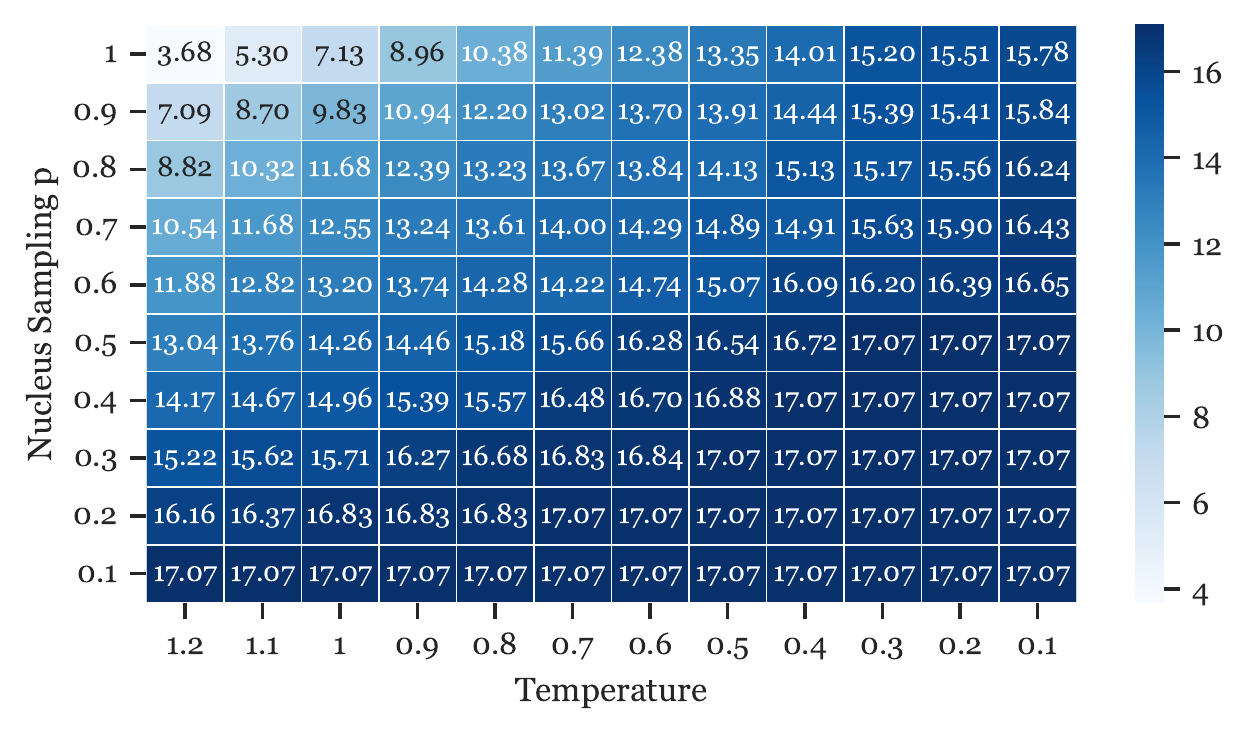}
        \caption{\textsc{pass@$1$}}
        \label{fig:param_search_1}
    \end{subfigure}\hfill%
    \begin{subfigure}[b]{0.5\linewidth}
        \centering
        \includegraphics[width=\linewidth]{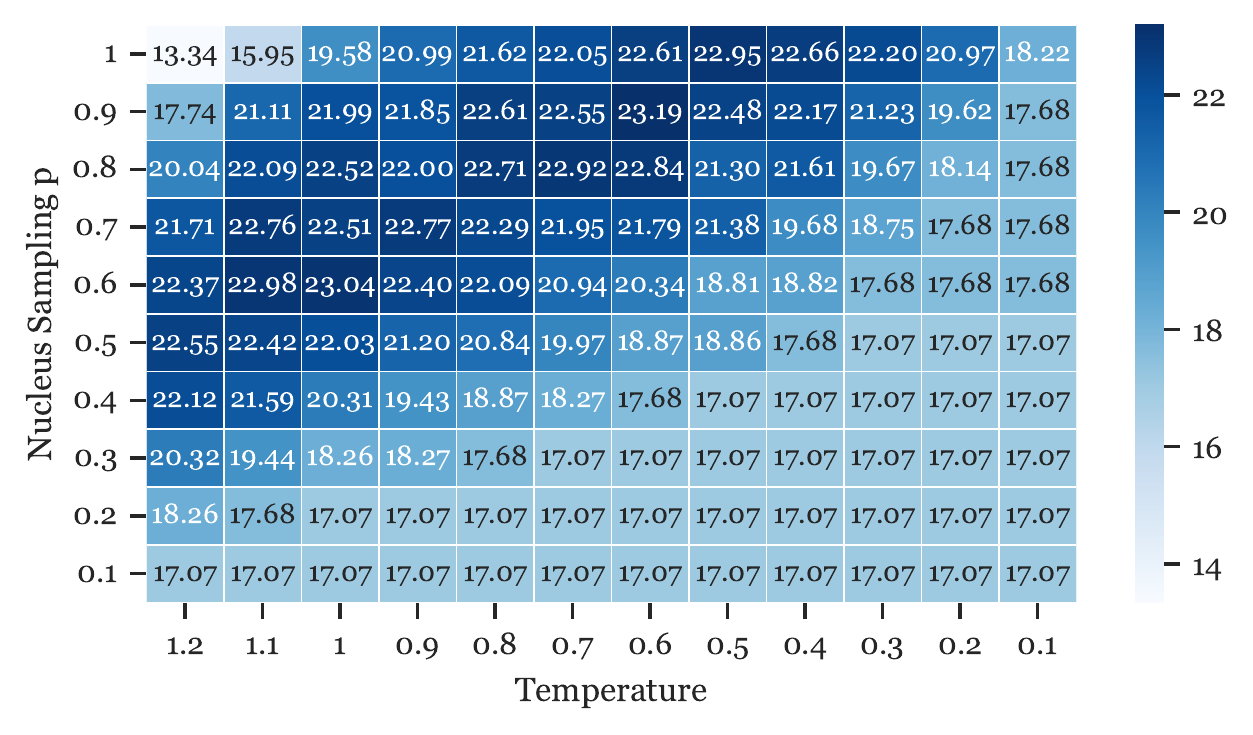}
        \caption{\textsc{pass@$10$}}
        \label{fig:param_search_10}
    \end{subfigure}
    \caption{Effect of temperature scaling and nucleus sampling on pass@$1$ and pass@$10$ on HumanEval.}
    \label{fig:param_search}
\end{figure}

Figure~\ref{fig:param_search} shows a study on the effect of temperature scaling and nucleus sampling on the best checkpoint of \codePLM~trained via the \textsc{Code-CLM} training objective. This study is focused on the HumanEval dataset and we gather $n=50$ samples per problem to calculate pass@$1$ and pass@$10$. We observe similar findings as observed by \citet{codex}, with different decoding strategies being better suited for different values of $k$ in pass@$k$, e.g. higher precision (low $p$ and temperature) is required for low values of $k$ while higher recall (high $p$ and temperature) improves larger $k$ pass ratios. 

An interesting observation is that the hyper-parameter values that are optimal for pass@$1$ performance are essentially equivalent to performing \textit{greedy decoding}. As such, we use greedy decoding to evaluate performance of various checkpoints over training as it is much more efficient.

\subsubsection{Impact of Different Prompting Strategies}
\label{sec:prompt_effect}

As shown in Table~\ref{tab:HE_example}, the HumanEval dataset problems also provide some input unit tests, and the provided function signatures may contain argument and return value typing. Specifically, unit tests are available for 136 of 164 problems, while function type declarations are present in 30 of 164 problems. We note that we are exclusively examining the unit tests that are available as part of HumanEval's prompt and not the held-out unit tests used for evaluation of the generated solutions. By default, the dataset prompt is formed to incorporate this additional information as shown in the example below. 

\begin{lstlisting}[style=custom]
<descr> Return a greatest common divisor of two integers a and b. \n >>> greatest_common_divisor(3, 5) \n 1 \n >>> greatest_common_divisor(25, 15) \n 5 <python>\n def greatest_common_divisor(a: int, b: int) -> int:\n
\end{lstlisting}

However, such information is usually absent in data extracted from online code repositories, such as those \codePLM~is trained on. To bring the HumanEval prompts closer to the format the model was exposed to during training, we explored some variations by removing the unit tests and/or typing. Both removals are performed automatically via simple heuristics.

\begin{lstlisting}[style=custom]
<descr> Return a greatest common divisor of two integers a and b. <python>\n def greatest_common_divisor(a, b)\n
\end{lstlisting}

Table~\ref{tab:ut_type} shows pass@$k$ results with different prompt variations on the best checkpoint of \codePLM~trained via the \textsc{Code-CLM} training objective. We observe some improvement on pass@$10$ performance when excluding both unit-tests and typing from the prompt, but no consistently significant differences altogether. Similar behavior is observed on our fine-tuned models (Section~\ref{sec:FT-results}).

\begin{table}[h]
\centering
    \begin{tabular}{ccccc}
        \toprule
        Unit-Tests & Typing &\textsc{pass@$1$} & \textsc{pass@$10$} & \textsc{pass@$100$} \\
        \cmidrule(lr){1-2} \cmidrule(lr){3-5}
        \cmark & \cmark & \textbf{17.07} & 23.02 & \textbf{34.36}\\
        \cmark & \xmark & 15.24 & 23.05 & 34.21\\
        \xmark & \cmark & \textbf{17.07} & 24.01 & 33.29\\
        \xmark & \xmark & 16.46 & \textbf{24.05} & 33.08\\
        \bottomrule \\
    
    \end{tabular}
    \caption{Effect of including unit tests and typing in the prompt of the HumanEval dataset.}
\label{tab:ut_type}
\end{table}




\subsubsection{Impact of Pre-training Strategies}

We also test the effect that each pre-training objective has on the model performance for the HumanEval dataset in Table \ref{tab:training_ablation}.
\begin{table}[h!]
    \centering
    \begin{tabular}{llrrr}
    \toprule
        \multicolumn{2}{c}{\textsc{Objective}}
        & \textsc{pass@$1$} & \textsc{pass@$10$} & \textsc{pass@$100$}\\ \cmidrule(lr){1-2} \cmidrule(lr){3-5} 
        Stage-1 & \textsc{CLM}         & 10.37 &	13.95 &	25.32 \\ \cmidrule(lr){2-5}
        \multirow{3}{*}{Stage-2} 
        & \textsc{Code-CLM}    & \textbf{17.07}  & \textbf{24.05} & \textbf{34.55} \\
        & \textsc{Docstr-MLM + Code-CLM}  & 12.80  & 15.27 &	24.16 \\ 
        & \textsc{Docstr-MCLM + Code-CLM} & 12.19  & 16.48 &	24.92 \\
    \bottomrule \\
    \end{tabular}
    \caption{Comparison of different training strategies on the HumanEval dataset for the 317M model.}
    \label{tab:training_ablation}
\end{table}
\vspace{-0.5cm}

We observe that objectives that calculate a loss on the docstring (i.e., \textsc{Docstr-MCLM} and \textsc{Docstr-MLM}) perform poorly compared to training the model with loss only on the code-tokens (\textsc{Code-CLM}).
We attribute this phenomenon to the fact that in these objectives, the docstring token embeddings receive a joint update via backpropagation combining the loss on the code tokens and the loss on the docstring tokens. 
As a result, the models try to optimize at the same time two conflicting objectives, i.e. updates that lead to better docstring generation/understanding and updates leading to better code generation.
On the contrary, when there is no loss computed over the docstring, the token embeddings are updated solely based on what will result in better code generation and consequently give higher performance.
Compared with stage-1, all objectives manage to improve during stage-2 across pass@$1$ and pass@$10$, while those using docstring-losses seem to hurt pass@$100$. Generally, we observe that all objectives lead to higher gains for lower $k$ values of pass@$k$. This implies that the objectives overfit on single correct solutions per problem, leading to a prioritization of precision over recall. 

\begin{figure}[!h]
    \centering
    \begin{subfigure}[b]{0.33\linewidth}
        \includegraphics[width=\linewidth]{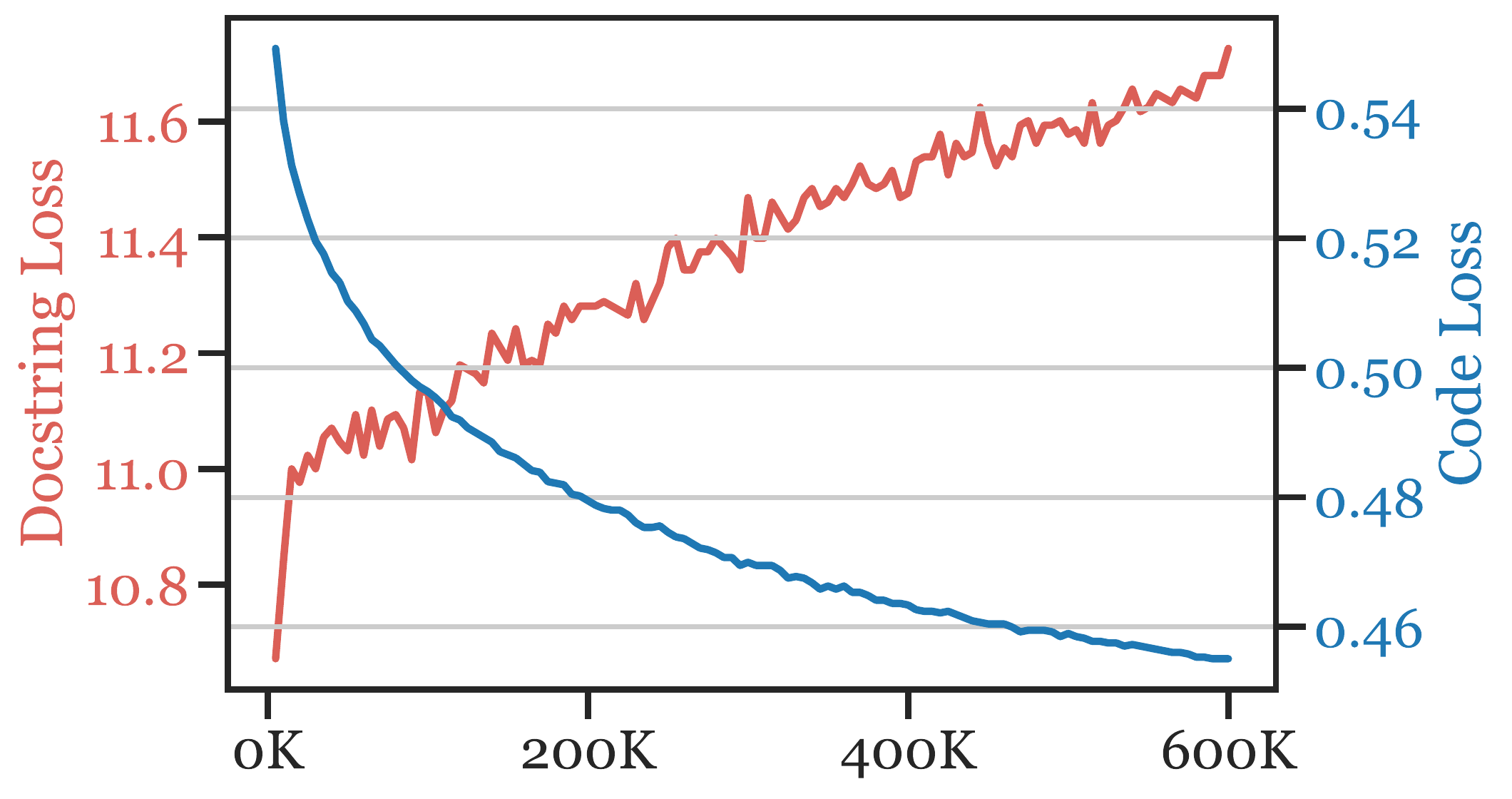}
        \caption{\textsc{Code-CLM}}
        \label{fig:code_clm_losses}
    \end{subfigure}%
    \begin{subfigure}[b]{0.33\linewidth}
        \includegraphics[width=\linewidth]{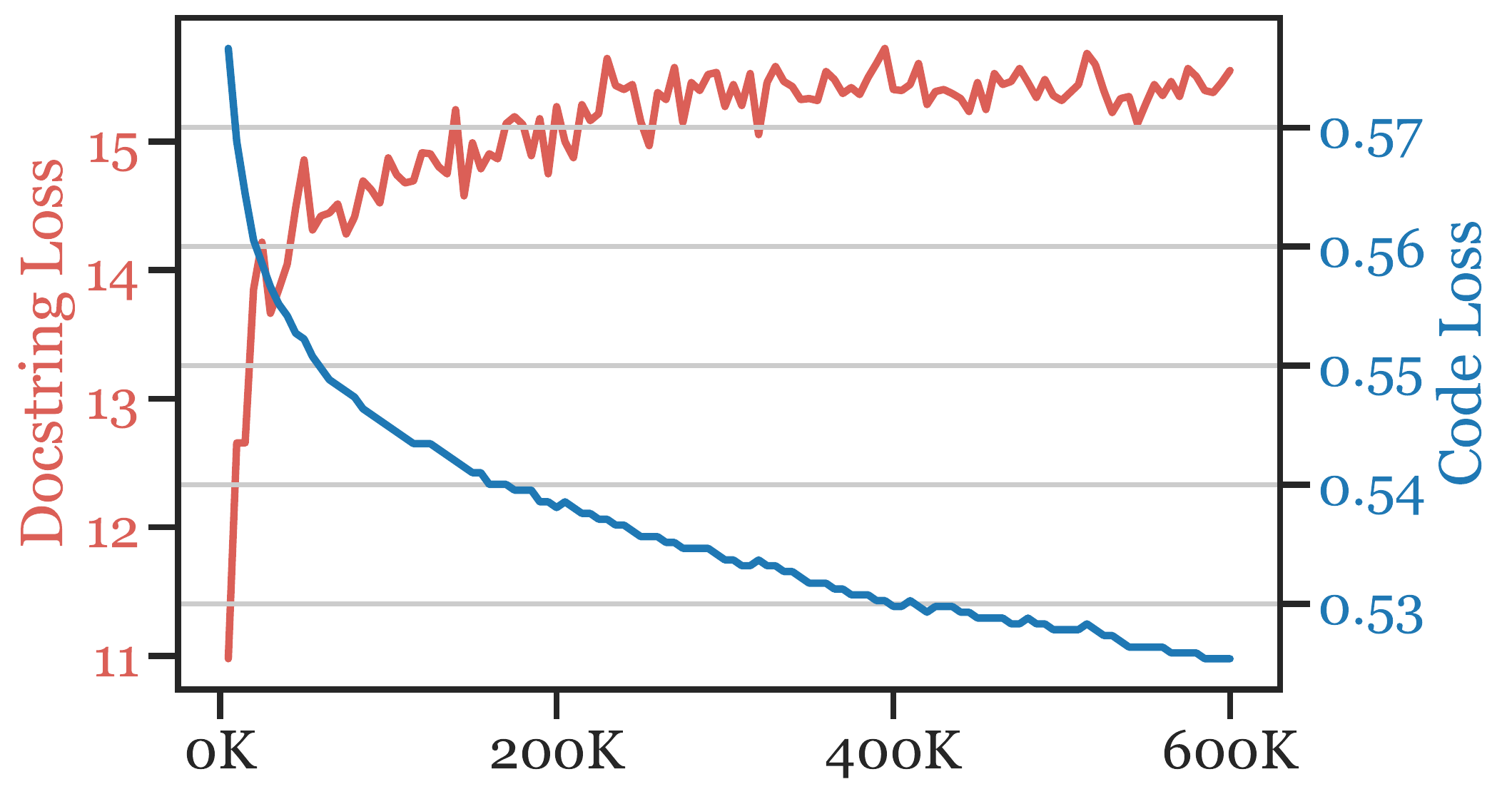}
        \caption{\textsc{Docstr-MLM}}
        \label{fig:docstr_mlm_losses}
    \end{subfigure}%
    \begin{subfigure}[b]{0.33\linewidth}
        \includegraphics[width=\linewidth]{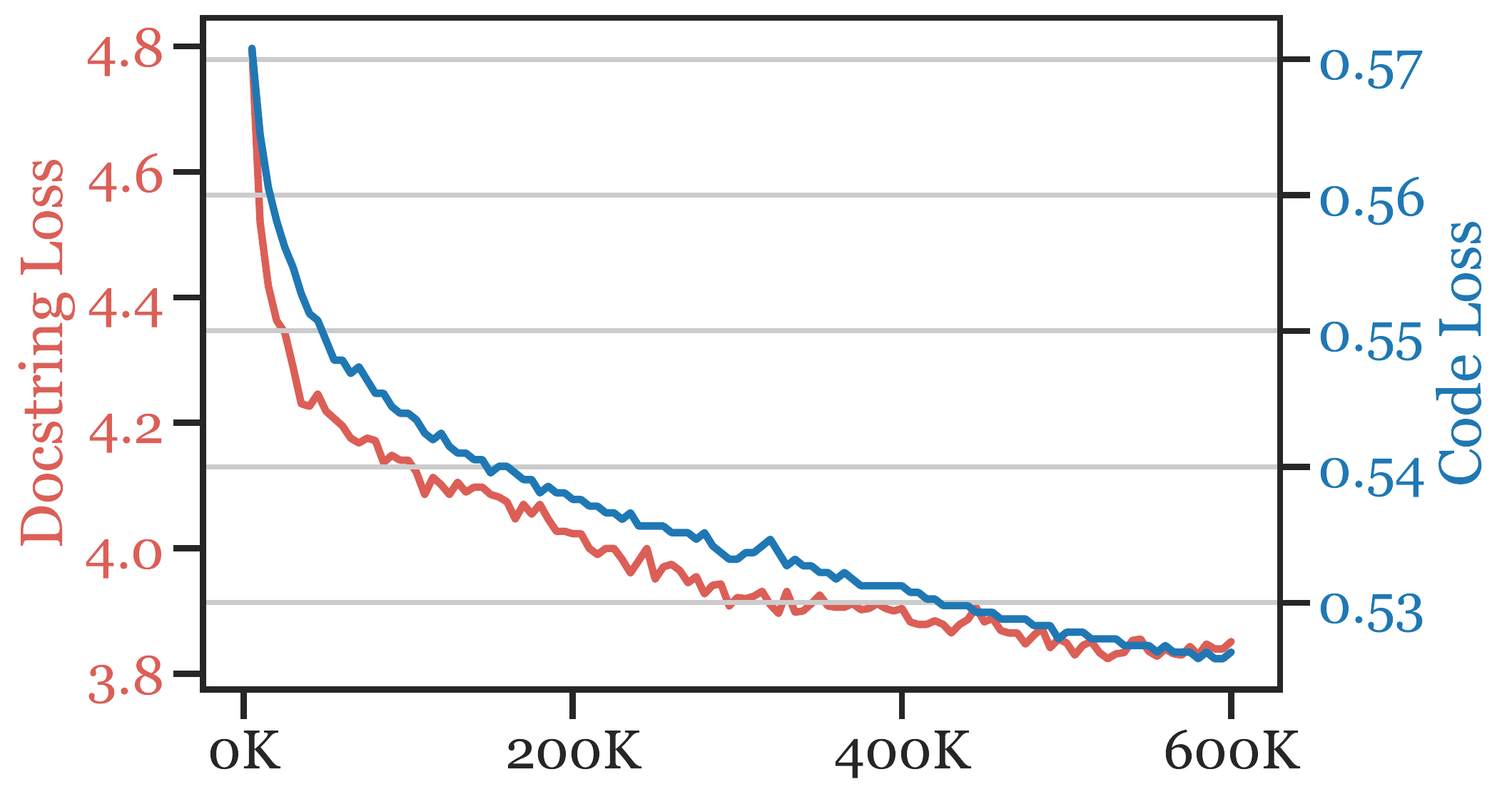}
        \caption{\textsc{Docstr-MCLM}}
        \label{fig:docstr_mclm_losses}
    \end{subfigure}
    \caption{Validation set Docstring and Code causal losses when employing different objectives on the 317M model.}
    \label{fig:docstr_code_losses}
\end{figure}

To further analyse the impact of the objectives on the two subsequences, we plot the causal loss (i.e., next token prediction) on the docstring (docstring loss) and code tokens (code loss) separately for each of these objectives as shown in Figure \ref{fig:docstr_code_losses}. 
We opt to examine the CLM loss rather than any alternatives, to determine whether there is any catastrophic forgetting happening as a result of abandoning the CLM loss on docstring as it was used in stage-1.

The \textsc{Code-CLM} objective achieves lower code loss compared to the other objectives (0.46 versus 0.53), further supporting its improvements. Interestingly, the code loss on the other objectives also decreases but the conflicting objectives do not allow it to achieve \textsc{Code-CLM} performance.
Regarding docstring loss, we observe that for \textsc{Code-CLM} and \textsc{Docstr-MLM} the  loss keeps increasing over time, while for the \textsc{Docstr-MCLM} objective, both losses decrease over the course of training. This makes sense, intuitively, as both \textsc{Code-CLM} and \textsc{Docstr-MLM} abandon CLM on docstring. For \textsc{Docstr-MCLM}, even though the loss on the docstring is decreased, this does not lead to lower code loss and consequently fails to increase performance, indicating an interesting direction of how we can best simultaneously encode natural language and programming language.


\subsubsection{Impact of Shared Embeddings}
\label{sec:shared_embed}

We check the impact of having separate embeddings between the code and the docstring on our best objective \textsc{Code-CLM}, as discussed in Section \ref{sec:tokenization}. Table \ref{tab:tokenization_ablation} illustrates an increasing drop in performance for different values of $k$, with $-1.2$ percentage points on pass@$1$, $-2$ on pass@10 and $-2.7$ on pass@100, when the token embeddings between code and docstring are shared.
This finding indicates that keeping separate embeddings between the two inputs can be beneficial for the task.
\begin{table}[h!]
    \centering
    \begin{tabular}{crrr}
    \toprule
        \textsc{Docstr-Code Embeddings}
        & \textsc{pass@$1$} & \textsc{pass@$10$} & \textsc{pass@$100$}\\ \cmidrule(lr){1-1} \cmidrule(lr){2-4}
        
        separate &  17.07 & 24.05 & 34.55 \\
        shared   &  15.85 & 22.01 & 31.82 \\ 
    \bottomrule \\
    \end{tabular}
    \caption{Comparison between different tokenizations between docstring and code for the \textsc{Code-CLM} objective on the HumanEval dataset for the 317M model.}
    \label{tab:tokenization_ablation}
    \vspace{-5mm}
\end{table}

Enabling shared embeddings enforces an explicit connection between docstring and code tokens in terms of their semantics. Although this can be true for some natural language-like programming languages, as in our case Python, keeping separate embeddings can assist learning of docstring token representations for better code generation. 

\section{Fine-Tuning Methodology}
\label{finetuning}

In this section, we evaluate a version of \codePLM~that was fine-tuned on a combination of competitive programming problems and code from continuous integration tests. The model, called \codePLM-FT, allows us to estimate the improvement in problem solve rates given additional data from a more similar domain / distribution, which we describe in the following paragraphs. 

\paragraph{Competitive Programming Data}
\label{sec:competitive}
The following two datasets provide dozens of successful code completions for each problem. Therefore, we can upsample each dataset by pairing up to five solutions with each problem description to avail the model of different correct solutions for the same problem specification. APPS~\cite{hendycks-etal-2021-apps} was proposed to benchmark the programming capability of code-generating models. It includes 10,000 programming tasks (5,000 train, 5,000 test) that require code generation / completion given a detailed problem description\footnote{\url{https://github.com/hendrycks/apps}}. Program arguments are provided from standard input as well as function calls, sometimes embedded inside the class. We use the train and test set programs, which we filter for the maximum input length of the model (1,024), retaining 43K examples. Code Contests (CC) is a similar dataset~\cite{alpha_code} for training and evaluation of competition-level code generation\footnote{\url{https://github.com/deepmind/code_contests}}, containing over 13K programming problems. After upsampling and length filtering, we retain 18K instances for fine-tuning.

\paragraph{Continuous Integration Data} 

In addition to the competitive programming data, we also gathered a dataset consisting of \textit{correctly-implemented} functions from open source projects. Following recent work on evaluating language models trained on code \cite{codex}, we considered public Python repositories on GitHub using Travis\footnote{\url{https://www.travis-ci.com/}} or Tox\footnote{\url{https://tox.wiki/en/latest/}} as their continuous integration (CI) system(s). First, we reproduced the CI environment of each project locally in docker instances, then injected a tracing function in the environment, which would capture the input, output, invocation location and context code for each function invoked during integration testing. After that, the CI script was triggered and the data of invoked functions were recorded into a database. We further clustered the collected data into 4 groups, according to the external contextual information (dependencies) that a function requires to run. That is, functions depending on Python-builtin objects, Python-standard libraries, Pypi-public libraries, and the residing class. We expect this variety could enable the generation of different function types, e.g., self-contained, module, member or class functions. After filtering for maximum model input length, 49K examples are retained for fine-tuning.

\paragraph{Training Details} We fine-tune \codePLM-FT for 5 epochs on the aforementioned data, using the \textsc{Code-CLM} objective. We decrease the batch size to 32 and use a linear decay learning rate scheduler. All other model training and program generation settings follow the Stage-2 protocol (Section \ref{sec:pretraining-results-analysis}).

\subsection{Fine-tuning Results}
\label{sec:FT-results}

\codePLM-FT clearly benefits from additional fine-tuning on data that is closer to the target distribution, i.e. short programming tasks solving competitive and technical interview questions (Table \ref{tab:palms}). The pass rate improvement is more pronounced for MBPP, which may be due to the lower difficulty of problems (relative to HumanEval). Recall that the B in MBPP stands for 'basic'. In the following paragraphs, we discuss the importance of appropriate fine-tuning data and provide additional methods for boosting the pass rates of solutions generated by \codePLM(-FT).

\subsubsection{Impact of In-domain Data}
MBPP and HumanEval provide a challenging zero-shot evaluation of models designed to generate functionally correct programs. MBPP\footnote{\url{https://github.com/google-research/google-research/tree/master/mbpp}} additionally provides a small training/prompting set of 474 instances, which we can use to contrast the rate of improvement between out-of-domain and in-domain data. Fine-tuning \codePLM~on just a few hundred relevant problems increases pass@$k$ almost as much as tens of thousands of correctly implemented but mostly out-of-domain examples (see Table \ref{tab:palms}). Models therefore need to acquire a wide spectrum of problem solving and programming knowledge to perform well on these datasets. Note that despite their superficial similarity, MBPP fine-tuning provides no benefit to HumanEval (Table \ref{tab:palms}). 

\begin{table}[h!]
\centering
\scalebox{0.95}{
\begin{tabular}{lccc|ccc}
    \toprule
    \multirow{2}{*}{\textsc{Models}} & 
    \multicolumn{3}{c|}{\textsc{MBPP (\%)}} & 
    \multicolumn{3}{c}{\textsc{HumanEval (\%)}} \\ 
    & \textsc{pass@$1$} & \textsc{pass@$10$} & \textsc{pass@$100$} 
    & \textsc{pass@$1$} & \textsc{pass@$10$} & \textsc{pass@$100$} \\  
    \cmidrule(lr){1-1} \cmidrule(lr){2-4} \cmidrule(lr){5-7}
    
    \textsc{\codePLM} & 16.20 & 34.39 & 53.74 & 17.07 & 24.05 & 34.55 \\ 
    \textsc{\codePLM}-FT & 24.60 & \textbf{44.19} & \textbf{63.07} & \textbf{19.50} & \textbf{25.96} & \textbf{40.80} \\
    \codePLM-MBPP & \textbf{25.40} & 43.32 & 60.03 & 15.24 & 22.73 & 32.65 \\ \bottomrule \\
\end{tabular}
}
\caption{\codePLM-FT (fine-tuned with competitive and continuous integration data), \codePLM-MBPP (fine-tuned on MBPP train set only), \codePLM (trained on open-source data).}
\vspace{-5mm}
\label{tab:palms}
\end{table}


\subsubsection{Impact of Unit Tests}
HumanEval provides input/output pairs in the problem description that can be used to verify the correctness of a solution, e.g. "\texttt{remove\_letters(PHP,\ P) == H}". MBPP problem descriptions can be augmented with test cases provided for evaluation in order to investigate whether the model benefits from the additional insights. Table \ref{tab:asserts} shows the pass rates for models trained/fine-tuned (\textsc{Train}) with and without unit tests as well as programs generated (\textsc{Sample}) with and without asserts. We observe that for MBPP, the pass rates substantially increase with unit tests in the prompt, particularly when this is provided during sampling as well. This is not the case for HumanEval, however, possibly because the problem descriptions are longer, more detailed and less formulaic. Similar observations were made during the zero-shot analysis in Section~\ref{sec:prompt_effect}. A more effective way of improving HumanEval scores is to use the same unit tests to filter out dysfunctional programs with our proposed methods, outlined in section \ref{sec:filtering}.

\begin{table}[h!]
\scalebox{0.87}{
\begin{tabular}{lccccc|ccc}
\toprule
\multirow{2}{*}{\textsc{Model}} & 
\multirow{2}{*}{\textsc{Train}} & 
\multirow{2}{*}{\textsc{Sample}} & 
\multicolumn{3}{c|}{\textsc{MBPP (\%)}} & 
\multicolumn{3}{c}{\textsc{HumanEval (\%)}} \\ 
& & & \small\textsc{pass@$1$} & \small\textsc{pass@$10$} & \small\textsc{pass@$100$} &  \small\textsc{pass@$1$} & \small\textsc{pass@$10$} & \small\textsc{pass@$100$} \\ \cmidrule(lr){1-3} \cmidrule(lr){4-6} \cmidrule(lr){7-9}

\codePLM & N/A & \cmark & 16.20 & \textbf{34.39} & \textbf{53.74} & \textbf{17.07} & \textbf{24.05} & \textbf{34.55} \\
\codePLM & N/A & \xmark & \textbf{17.40} & 34.29 & 50.91 & 16.46 & 23.38 & 33.58 \\ \midrule
\hspace{0.15cm} + MBPP-train & \xmark & \xmark & 20.00 & 37.70 & 53.04 & 15.85 & \textbf{24.11} & \textbf{33.66} \\
\hspace{0.15cm} + MBPP-train & \xmark & \cmark & 23.60 & 40.91 & 58.91 & 13.41 & 23.52 & 33.44 \\
\hspace{0.15cm} + MBPP-train & \cmark & \xmark & 21.00 & 36.44 & 51.92 & \textbf{16.46} & 23.79 & 31.50 \\
\hspace{0.15cm} + MBPP-train & \cmark & \cmark & \textbf{25.40} & \textbf{43.32} & \textbf{60.03} & 15.24 & 22.73 & 32.65 \\ \bottomrule \\
\end{tabular}
}
\caption{\codePLM~trained and/or solutions generated with and without unit tests.}
\label{tab:asserts}
\vspace{-5mm}
\end{table}

\subsubsection{Filtering Generated Programs}
\label{sec:filtering}

In order to continue to improve \codePLM's pass rates, we can follow the program generation stage with an additional post-processing step, e.g. filtering on unit tests, declared function types and checking for syntactic correctness. The solutions ($n=200$) to each HumanEval problem are filtered before being evaluated on the held-out unit tests. Not all problems have embedded/parsable unit tests or have declared all function arguments and return types. Therefore, we evaluate the effectiveness of our filtering methods on problem subsets that have the specific filter available, i.e. 136 of 164 with unit tests and 30 of 164 with full type declarations (all 164 problems can be checked for syntax errors). This gives us a baseline score when no filtering is applied. The number of problems with each filter and the corresponding pass rates are shown in Table \ref{tab:filter}.

\begin{table}[h!]
\centering
\begin{tabular}{llcccc}
\toprule
\multirow{2}{*}{\textsc{Filtering Method}} & & 
\multicolumn{2}{c}{\codePLM} & 
\multicolumn{2}{c}{\codePLM-FT} \\
& & \textsc{pass}@1 & \textsc{pass}@10 
& \textsc{pass}@1 & \textsc{pass}10 \\ \midrule

\multirow{2}{*}{\textsc{Unit Testing}} 
& \textsc{Base} & 14.53 & - & 16.67 & - \\
& \textsc{Filter} & \textbf{35.48} & - & \textbf{41.52} & - \\ \cmidrule{2-6}

\multirow{2}{*}{\textsc{Typing}}
& \textsc{Base} & 25.16 & 50.65 & 30.45 & 54.61 \\
& \textsc{Filter} & \textbf{26.99} & \textbf{52.00} & \textbf{31.30} & \textbf{55.72} \\ \cmidrule{2-6}

\multirow{2}{*}{\textsc{Invalid Syntax}} 
& \textsc{Base} & 12.05 & 23.27 & 13.85 & 25.40 \\
& \textsc{Filter} & \textbf{12.06} & 23.27 & \textbf{13.86} & \textbf{25.44} \\

\bottomrule \\

\end{tabular}
\caption{HumanEval pass@$k$ with/out filtering (for each problem subset). Unit tests are available for 136/164 problems, argument/return types for 30/164 problems and syntax check for all problems.}
\label{tab:filter}
\vspace{-5mm}
\end{table}

\paragraph{Unit Tests}

In many real-world scenarios, the description of a problem can include example function input(s) and output(s), which we can use to filter out unsuccessful solutions, potentially increasing the overall pass rate. In addition to the held-out tests, HumanEval typically provides unit tests embedded in the description, expressed in a natural way and usually following a simple pattern. A basic regular expression (to avoid overfitting) is used to extract the majority (136/164) of the unit tests. Table \ref{tab:filter} shows a significantly improved pass rate after this filtering step. Only the pass@$1$ score is available with this filter, as many problems fail to reach at least 10 solutions per problem, once filtered. The mean/median number of solutions after filtering is 115/155, an average reduction of around 40-45\%.

\paragraph{Typing}

In other scenarios, we may be provided with function argument types and the expected return type. In such cases, we can generate plausible function inputs, execute the sampled program and evaluate the type of the return value. If the program returns the expected type, it will be included for the final pass@$k$ evaluation. As the generated programs are executed, we implicitly verify the syntactic and semantic correctness, too. Relying only on checking the return type will inevitably lead to false positives as programs often return the correct type, e.g. a float or a list of floats but with the wrong value(s). However, even this simple filter can improve the final pass rates (Table \ref{tab:filter}) and reduce the number of programs by around 25\% (mean/median 150/152 after filtering). 

\paragraph{Invalid Syntax}

In the absence of unit tests and input/output type declarations retrieved from the description, filtering out syntactically invalid solutions can be a fast and simple method for eliminating incorrect problem solutions. With that in mind, Table \ref{tab:filter} shows that \codePLM~is generating syntactically well-formed solutions almost all the time hence we filter $<1$ out of 200 problem solutions (mean/median after filtering 199/200), leaving the pass rates unchanged.

\subsection{Data Selection with Few-Shot Similarity}

Transformer-based pretrained language models continue to rapidly increase in size \cite{wang-etal-2021-codet5,alpha_code,google_palm} hence it is desirable to reduce the compute requirements while maintaining good performance, whenever possible. To this end, we simulate a scenario, in which we possess a small sample of programming problems (possibly written from scratch) and a much larger dataset of heterogeneous training data. We then experiment with choosing training examples from the much larger dataset based on their similarity to a centroid embedding that represents our 'few-shot' sample. We randomly select 10 problems (description and code) from HumanEval and 10 from MBPP (to avoid overfitting to HumanEval) as our sample, denoted $P = [p_{0}, p_{1}, .., p_{20}]$ in Equation \ref{eq:proto}. The centroid embedding $p_{emb}$ is obtained using a version of CodeBERT~\cite{feng-etal-2020-codebert}, which has been further trained on CodeSearchNet \cite{DBLP:journals/corr/abs-1909-09436} using the Replaced Token Detection loss \cite{clark-etal-2020-electra}. The model is denoted by $f$ in Equations \ref{eq:proto} and \ref{eq:dist}.
\begin{gather}
\label{eq:proto}
    p_{emb} = \frac{1}{|P|}\sum_{i=1}^{P}f(p_i) \\ \vspace{1.5cm}
\label{eq:dist}
    distance(x_i) = MSE \left( f(x_i),\ p_{emb} \right) 
\end{gather}

\begin{figure}[t!]
    \centering
    \begin{subfigure}[b]{0.33\linewidth}
        \includegraphics[width=\linewidth]{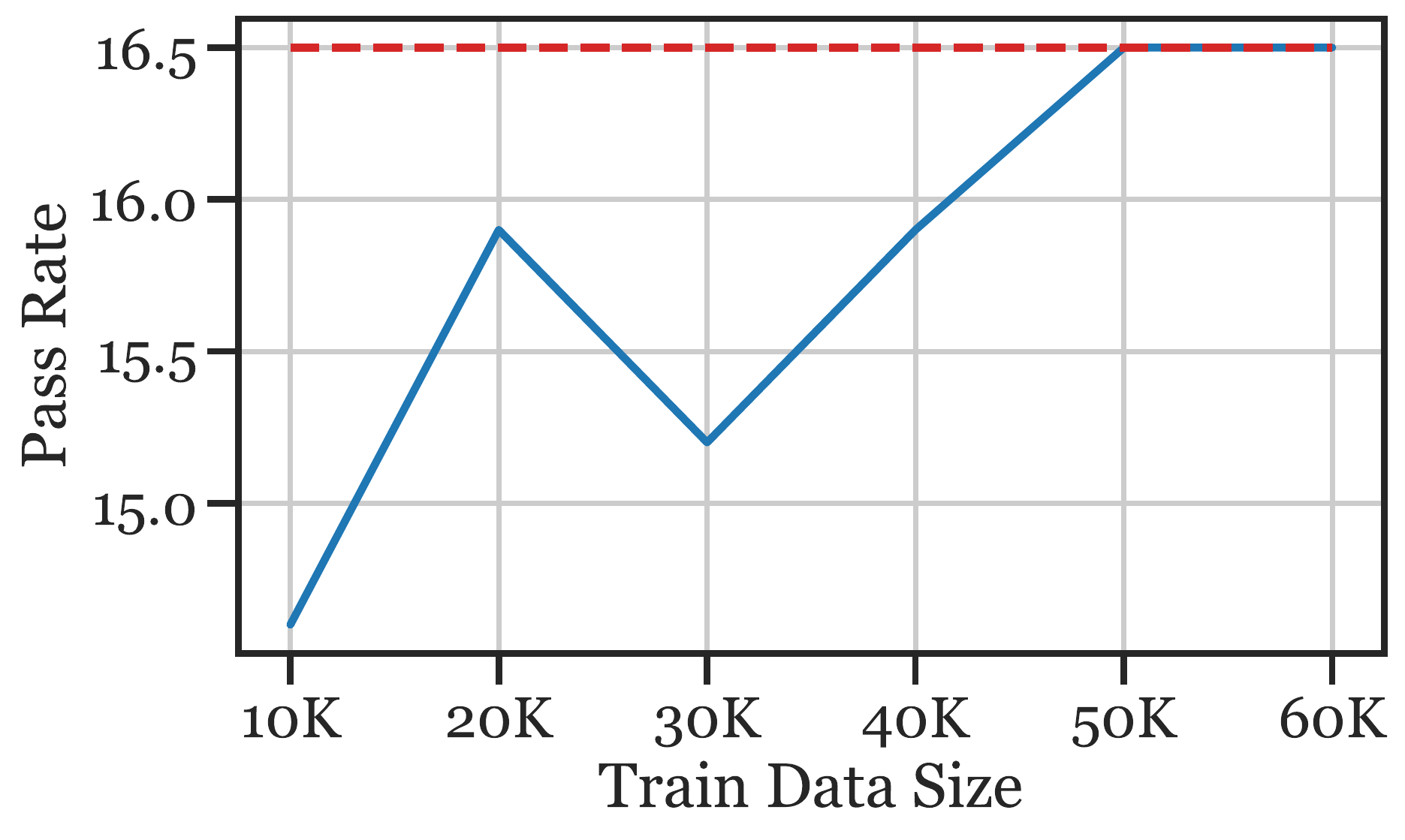}
        \caption{\textsc{pass@$1$}}
        \label{}
    \end{subfigure}\hfill%
    \begin{subfigure}[b]{0.33\linewidth}
        \includegraphics[width=\linewidth]{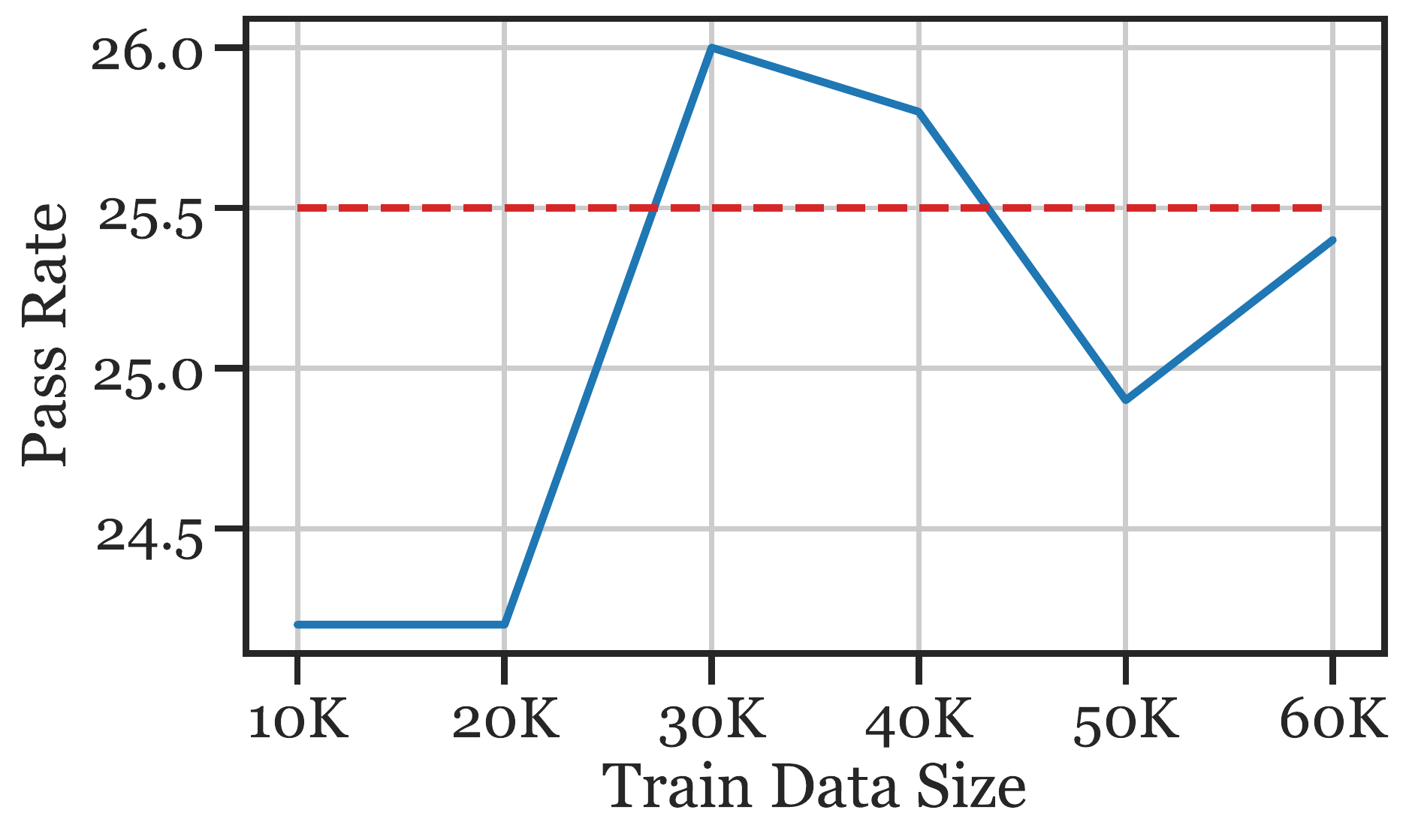}
        \caption{\textsc{pass@$10$}}
        \label{}
    \end{subfigure}\hfill%
    \begin{subfigure}[b]{0.33\linewidth}
        \includegraphics[width=\linewidth]{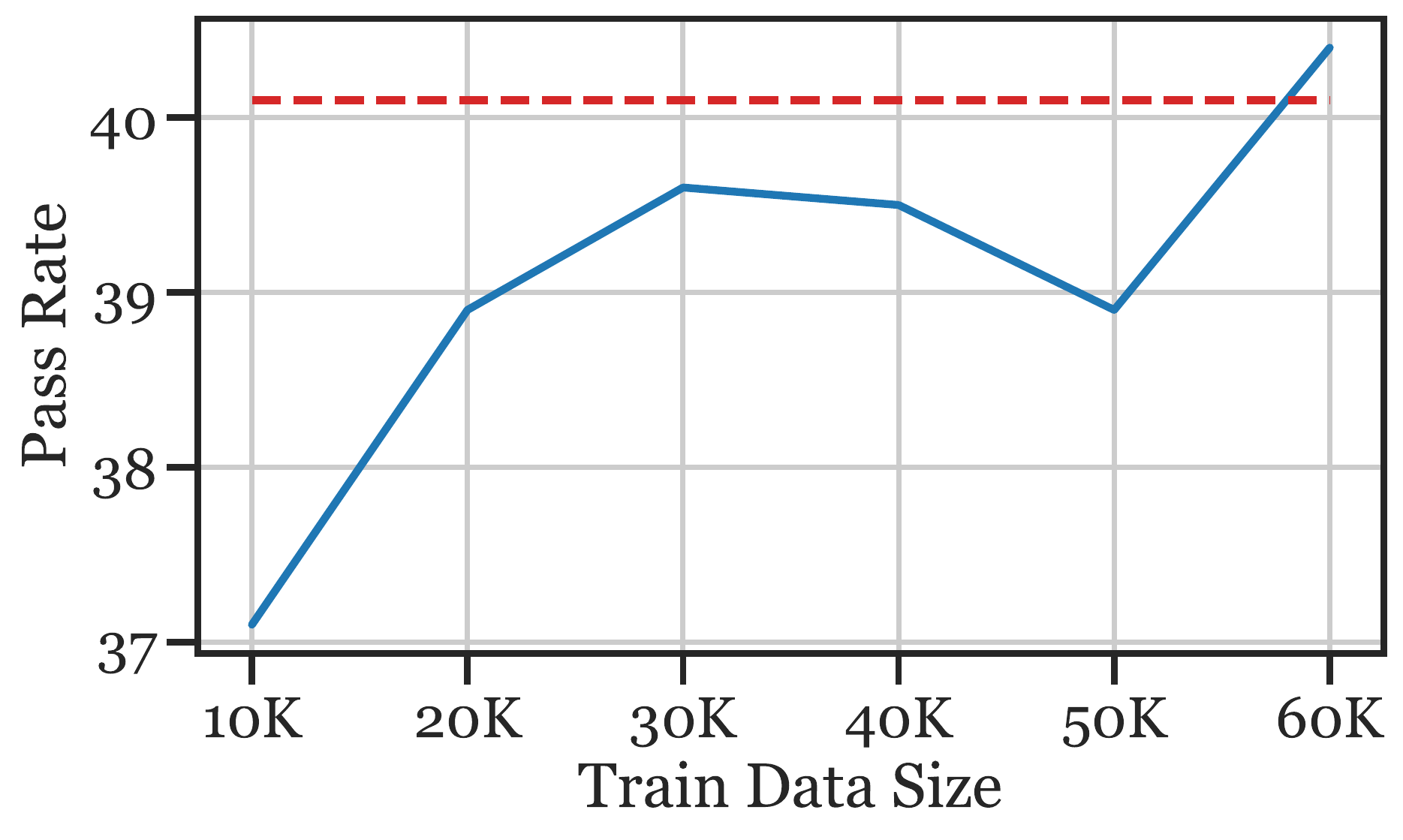}
        \caption{\textsc{pass@$100$}}
        \label{}
    \end{subfigure}
    \caption{The pass rates as we increase the fine-tuning data. Each time, 10K nearest examples are added. The red line shows the pass@$k$ when \codePLM~is fine-tuned with all 120K examples.}
    \label{fig:similarity}
\end{figure}

A training example $x_i$ is ranked/chosen by its Mean Squared Error (MSE) $distance$ (lowest first) from $p_{emb}$. \codePLM~is then fine-tuned for five epochs with training data incremented by 10K, up to 60K examples, which is 50\% of the total data. The pass@k are shown in Figure \ref{fig:similarity}.

Perhaps unsurprisingly, the competitive programming share of the fine-tuning data increases from 46.4\% (in the full 120K dataset) to 68.7\% after our similarity-based selection. We observe that for all pass@$k$, the full (dataset) score is reached or exceeded by fine-tuning on 25\% - 50\% of the larger dataset (see Figure \ref{fig:similarity}). Using this method, we are able to reduce model compute requirements while maintaining (even surpassing) the original pass rate. If we wish to fine-tune \codePLM~on some target data distribution in the future, we require a small ``few-shot'' sample of problems (possibly written from scratch) to effectively subsample our larger training dataset to achieve a good pass rate while reducing computational resources.

\section{Related Work}
\label{sec:related_work}

Recently, there has been an increasing interest in applying deep learning methods for NLP to code understanding and generation tasks.
Among others, this is evidenced by the introduction of new NLP venues, e.g. Natural Language Processing Advancements for Software Engineering (NLPaSE)~\cite{NLPaSE2021}, which had its second installment at APSEC 2021 and the 2nd International Workshop on Software Engineering Automation: A Natural Language Perspective (NLP-SEA 2021)~\cite{nlpsea2021} at ASE 2021. Other new venues include the 1st Workshop on Natural Language Processing for Programming (NLP4Prog 2021) at ACL 2021~\cite{nlp4prog-2021-natural}, the Deep Learning for Code (DL4C) Workshop at ICLR 2022~\cite{dl4c} and the 1st International Workshop on Natural Language-based Software Engineering (NLBSE 2022), co-located with ICSE 2022~\cite{nlbse2022}.

\subsection{Pre-trained Language Models for Programming Language}

From an NLP research perspective, the recent advancements in code understanding and generation have been mostly focused on revisiting proven effective natural language understanding (NLU) and generation (NLG) methods. CodeBERT~\cite{feng-etal-2020-codebert}, for instance, was trained using a combination of Masked Language Modeling inspired by \cite[BERT]{devlin-etal-2019-bert} and Replaced Token Detection \cite[ELECTRA]{clark-etal-2020-electra}. CodeT5~\cite{wang-etal-2021-codet5} and PYMT5~\cite{clement2020pymt5} were built on top of \cite[T5]{raffel-etal-2020-exploring} while UniXcoder~\cite{guo-etal-2022-unixcoder} was based on UniLM~\cite{dong-etal-2019-unilm}. The most notable works in the area are discussed next. We distinguish between models focused on \textit{code understanding}, where the goal is to learn contextual representations of source code, and \textit{code generation}, where the aim is to translate between different programming and / or natural languages as well as to perform code completion and / or code repair.


\paragraph{Code Understanding}
Research on this topic started with the seminal work of \citet{10.1145/3290353} who introduced \texttt{code2vec}\cite{10.1145/3290353}, a neural model for representing snippets of code as continuous distributed vectors, extending the idea from \citet[word2vec]{mikolov2013efficient}. Inspired by the introduction of BERT~\cite{devlin-etal-2019-bert}, \citet{kanade2020pretrained} proposed CuBERT (Code Understanding BERT), a natural adaptation of the model trained on source code. CodeBERT~\cite{feng-etal-2020-codebert} instead, used both source code and Natural Language (NL) with a discriminative objective, inspired by \cite[ELECTRA]{clark2020electra}. The model receives a masked NL-code pair and the generator then predicts the masked tokens (the prediction might be different from the original token). Subsequently, CodeBERT is trained to predict which tokens were replaced by the generator. \citet{guo2021graphcodebert} noted that previous pretrained models treat code snippets as sequences of tokens while ignoring the inherent structure of code. They presented GraphCodeBERT, which showed that incorporating the data flow, i.e. a semantic-level structure of code extracted from the abstract syntax tree (AST), leads to richer code representations. \citet[TREEBert]{jiang2021treebert} instead used the actual AST together with code snippets.

SynCoBERT~\cite{DBLP:journals/corr/abs-2108-04556} trained the model with natural language, code and ASTs, but updated the training objective to include identifier prediction and AST Edge Prediction. In addition, Contrastive Learning~\cite{he2020Moco,SIMCLR} has been used as a training regime to help associate snippets of code with syntactically diverse but equivalent programs as well as with corresponding natural language descriptions \cite[ContraCode]{jain-etal-2021-contrastive,DBLP:journals/corr/abs-2201-10005} or with different modalities \cite[Code-MVP]{https://doi.org/10.48550/arxiv.2205.02029}.

\paragraph{Code Generation} 
Earlier approaches for code generation were mostly limited to simplified versions of code completion or code translation tasks \cite{10.5555/2337223.2337322,raychev2014code}. However, recent work has been proposed to tackle more complex problems, e.g. text-to-code generation (program synthesis) and code summarization. \citet{DBLP:journals/corr/abs-2102-04664} introduced CodeGPT, an adaptation of GPT~\cite{radford2019language} applied to code, as a baseline for the CodeXGLUE benchmark. TransCoder~\cite{roziere2020unsupervised} presented a pre-trained language model specifically focused on unsupervised translation of code between different programming languages. PLBART~\cite{ahmad-etal-2021-unified} employed a denoising objective over code and natural language via token masking, token deletion, and token infilling as noising strategies. CoText~\cite{phan-etal-2021-cotext} was build on top of T5 with a special focus on multi-task learning over multiple programming languages. Another model variant was introduced by~\citet[InCoder]{fried2022incoder}. Instead of training to perform code/text generation in a single left-to-right pass, InCoder is also able to edit existing/partial programs via an infilling training objective. The training regime involved randomly replacing spans of code/comments with a placeholder and asking the model to generate the replaced lines. 

More recently, a number of large pretrained language models have been proposed, primarily focused on the task of text-to-functional-code generation. \citet{codex} introduced CodeX~\cite{codex}, a set of GPT-based language models trained on publicly available code from GitHub, up to 12B parameters in size. In addition, the authors fine-tuned their models using a set of training problems from competitive programming websites and repositories with continuous integration for better program synthesis, called Codex-S. Codex-D was fine-tuned on the reverse task of generating the program description given a particular function/code. 

\citet{alpha_code} introduced AlphaCode, a set of sequence-to-sequence models with up to 41B parameters, trained on data from programming competitions, e.g. Codeforces\footnote{\url{https://codeforces.com/}} (similar to Codex-S) as well as GitHub code in several programming languages. AlphaCode produces problem solutions by overgeneration-and-ranking, i.e. the model samples multiple solutions and uses filtering and clustering to determine the best ones. Finally, CodeGen~\cite{nijkamp2022conversational} was proposed as a conversational text-to-code approach using large language models with sizes of up to 16B parameters. CodeGen-NL was trained on The Pile~\cite{thepile}, which contains around 6GB of Python code. CodeGen-Multi was then further trained on BigQuery, which includes data from 6 different programming languages (C, C++, Go, Java, JavaScript and Python). A third model, CodeGen-Mono, builds on top of CodeGen-Multi and was additionally trained on Python-only data.

\subsection{Code Datasets and Evaluation}

There is a number of datasets proposed for code understanding and generation tasks, many of which have recently been included in the CodeXGLUE benchmark~\cite{DBLP:journals/corr/abs-2102-04664}. CodeXGLUE comprises 14 datasets across 11 tasks, including clone detection, defect detection, cloze test and code search for understanding as well as code completion, code translation, code summarization, code repair, code refinement, document translation and text-to-code generation. Among these datasets are CONCODE~\cite{iyer-etal-2018-mapping}, a Java-based dataset with more than 100,000 examples of Java classes built from public Github projects and containing environment information 
along with Javadocs and the corresponding code. 
CodeSearchNet~\cite{DBLP:journals/corr/abs-1909-09436} is a benchmark specifically designed to evaluate systems on semantic code search, i.e. a code retrieval task based on natural language queries. 

The above tasks have mostly adopted NLP evaluation metrics, for example, text-to-code generation still widely uses CodeBLUE~\cite{DBLP:journals/corr/abs-2009-10297} and exact match of the outputs. CodeBLUE is an extension of the standard n-gram overlap metric BLEU~\cite{papineni-etal-2002-bleu}, designed to improve the evaluation of generated code. More recently, evaluation datasets and metrics started focusing on the functional correctness of generated programs. \citet[Codex]{codex} introduced the HumanEval dataset (c.f. Section\ref{sec:eval_data}) and \citet{austin2021program} introduced the \textit{Mostly Basic Programming Problems}~\cite[MBPP]{austin2021program} dataset. Both datasets are relatively small, comprising of 164 and 974 instances where descriptions of programming problems and their corresponding coding solutions are included. These problems are then evaluated in terms of their behavior, instead of being treated as natural language (either syntantically or semantically). Specifically, the datasets provide unit tests that can be used to evaluate how close outputs are to what is expected.
The main difference between HumanEval and MBPP is that the former does not include a training set as it was meant to be used for zero-shot evaluation.
Other datasets, such as APPS~\cite{hendycks-etal-2021-apps}, containing 10,000 problems and Code Contests~\cite[CC]{alpha_code} (13,610 problems) (c.f. Section \ref{sec:competitive}), leverage data from existing online coding platforms, resulting in a significantly larger dataset size but also more noise.

\section{Conclusions}
\label{sec:conclusions}

In this report, we presented \codePLM, 
a pre-trained language model for the task of text-to-code generation. \codePLM~is based on the Pangu-$\alpha$ architecture and was initially pre-trained on raw natural language and programming data using the CLM objective, and subsequently trained specifically on pairs of docstrings and functions using combinations of \textsc{Code-CLM}, \textsc{Docstr-MLM}, and \textsc{Docstr-MCLM} training objectives. 
Zero-shot evaluation of \codePLM~on the HumanEval and MBPP datasets, designed to measure whether outputs comprise functionally correct programs, shows that this training can help reach equivalent or better performance than similarly sized models while using a smaller context window and less training data. Further analysis examined the impact of various decoding and sampling strategies, training objectives, and embedding sharing. 

\codePLM-FT demonstrated that the scores of the base model can be improved at a faster rate by curating data more closely related to our target task as the model is quite sensitive to mismatches and shifts in fine-tuning data distributions. If we possess a small set of example problems from the target distribution, this sensitivity can be somewhat mitigated by choosing a subset of our fine-tuning data based on the similarity to a centroid embedding. We also evaluated various post-processing methods, some of which allowed us to significantly improve pass rates by filtering out failing programs. 


\section{Acknowledgements}
The authors thank Wei Zhang, Jun Yao, Qian Zhao, Feng Xu, Zongwei Tan for their great support to this work.
The authors also thank Philip John Gorinski for his feedback on an early version of this report.

\bibliography{bibs/references, bibs/anthology}
\bibliographystyle{plainnat}

\newpage
\appendix

\section{Qualitative Analysis of Example Outputs}

We show some examples of code solutions generated by the different versions of \codePLM~after stage-1 pre-training (c.f. Section~\ref{sec:stage_1_training}), after stage-2 training with the \textsc{Code-CLM} objective (c.f. Section~\ref{sec:stage_2_training}), and after fine-tuning (c.f. Section~\ref{finetuning}).

In this first example from HumanEval in Table~\ref{tab:example_1}, we can see that both Stage-1 and Stage-2 models are attempting to generate solutions, but only the fine-tuned model produces a correct one.

\begin{table*}[h!]
\lstset{linewidth=9.6cm}
\centering
\begin{tabular}{>{\footnotesize\sc}l>{\footnotesize\sc}cl}
\toprule
Prompt & &
\begin{lstlisting}[style=examples]
Return a greatest common divisor of two integers a and b. 
>>> greatest_common_divisor(3, 5)
1 
>>> greatest_common_divisor(25, 15)
5 \end{lstlisting} \\ \cmidrule{3-3}

Stage-1 & \colorbox{red!30}{\xmark} &
\begin{lstlisting}[style=python]
def greatest_common_divisor(a: int, b: int) -> int:
    if a == 0:
        return b
    if b == 0:
        return a
    return greatest_common_divisor(a // b, b // a)    \end{lstlisting} \\ \cmidrule{3-3} 

Stage-2 & \colorbox{red!30}{\xmark} &
\begin{lstlisting}[style=python]
def greatest_common_divisor(a: int, b: int) -> int:
    if a == 0 or b == 0:
        return 0
    else:
        return a if a > b else b \end{lstlisting} \\ \cmidrule{3-3}
 
FT & \colorbox{green!25}{\cmark} &
\begin{lstlisting}[style=python]
def greatest_common_divisor(a: int, b: int) -> int:
    if a > b:
        a, b = b, a
    while b!= 0:
        a, b = b, a % b
    return a \end{lstlisting} \\
\bottomrule
\end{tabular}
\caption{Example 1 from Human Eval}
\label{tab:example_1}
\end{table*}

In the second example from HumanEval in Table~\ref{tab:example_2}, we observe a similar pattern with Stage-1 and Stage-2 failing to produce a solution, but the fine-tuned succeeds. Interestingly, the Stage-2 model did not even attempt to provide a solution. Given the complexity of the problem, we can assume that the fine-tuned version was able to solve it because it has encountered it (or one quite similar to it) in the fine-tuning data.

\begin{table*}[!h]
\lstset{linewidth=10.2cm}
\centering
\begin{tabular}{>{\footnotesize\sc}l>{\footnotesize\sc}cl}
\toprule
Prompt & &
\begin{lstlisting}[style=examples]
In this Kata, you have to sort an array of non-negative 
integers according to number of ones in their binary 
representation in ascending order. For similar number 
of ones, sort based on decimal value.

It must be implemented like this:
>>> sort_array([1, 5, 2, 3, 4]) == [1, 2, 3, 4, 5]
>>> sort_array([-2, -3, -4, -5, -6]) == [-6, -5, -4, -3, -2]
>>> sort_array([1, 0, 2, 3, 4]) [0, 1, 2, 3, 4] \end{lstlisting} \\ \cmidrule{3-3}

Stage-1 & \colorbox{red!30}{\xmark} &
\begin{lstlisting}[style=python]
def sort_array(arr):
    return sorted(arr)  \end{lstlisting} \\ \cmidrule{3-3} 

Stage-2 & \colorbox{red!30}{\xmark} &
\begin{lstlisting}[style=python]
def sort_array(arr):
    pass \end{lstlisting} \\ \cmidrule{3-3}
 
FT & \colorbox{green!25}{\cmark} &
\begin{lstlisting}[style=python]
def sort_array(arr):
    return sorted(arr, key=lambda x: (bin(x).count("1"), x)) \end{lstlisting} \\
\bottomrule
\end{tabular}
\caption{Example 2 from Human Eval}
\label{tab:example_2}
\end{table*}

In the third and final example from HumanEval in Table~\ref{tab:example_3}, Stage-2 managed to find a correct solution but Stage-1 and fine-tuned fails. The fine-tuned version actually generates its own docstring that is very similar with the prompt, indicating that it must have encountered the problem in the fine-tuning data as well. However, the subsequent generated solution is not adequate to solve the problem.

\begin{table*}[!h]
\lstset{linewidth=9.6cm}
\centering
\begin{tabular}{>{\footnotesize\sc}l>{\footnotesize\sc}cl}
\toprule
Prompt & &
\begin{lstlisting}[style=examples]
Given an array of integers nums, find the minimum sum of 
any non-empty sub-array of nums. \end{lstlisting} \\ \cmidrule{3-3}

Stage-1 & \colorbox{red!30}{\xmark} &
\begin{lstlisting}[style=python]
def minSubArraySum(nums):
    if len(nums) == 0:
        return 0
    min_sum = nums[0]
    for i in range(1, len(nums)):
        if nums[i] < min_sum:
            min_sum = nums[i]
    return min_sum \end{lstlisting} \\ \cmidrule{3-3} 

Stage-2 & \colorbox{green!25}{\cmark} &
\begin{lstlisting}[style=python]
def minSubArraySum(nums):
    if not nums:
        return 0
    min_sum = nums[0]
    current_sum = nums[0]
    for i in range(1, len(nums)):
        current_sum = min(current_sum + nums[i], nums[i])
        min_sum = min(min_sum, current_sum)
    return min_sum \end{lstlisting} \\ \cmidrule{3-3}
 
FT & \colorbox{red!30}{\xmark} &
\begin{lstlisting}[style=python]
def minSubArraySum(nums):
    """
    Given an array of integers nums, find the minimum sum 
    of any non-empty sub-array of nums. 
    Example minSubArraySum([2, 3, 4, 1, 2, 4]) == 1 
    minSubArraySum([-1, -2, -3]) == -6
    """
    min_sum = nums[0]
    for i in range(1, len(nums)):
        min_sum = min(min_sum + nums[i], nums[i])
    return min_sum \end{lstlisting} \\
\bottomrule
\end{tabular}
\caption{Example 3 from HumanEval}
\label{tab:example_3}
\end{table*}

We observe similar behavior in Tables~\ref{tab:example_1_mbpp} and~\ref{tab:example_2_mbpp} which show two corresponding examples for the MBPP dataset.

\begin{table*}[!h]
\lstset{linewidth=8.3cm}
\centering
\begin{tabular}{>{\footnotesize\sc}l>{\footnotesize\sc}cl}
\toprule
Prompt & &
\begin{lstlisting}[style=examples]
Write a function to find whether a given array of 
integers contains any duplicate element. 
Your code should pass these tests:
assert test_duplicate(([1,2,3,4,5]))==False
assert test_duplicate(([1,2,3,4, 4]))==True
assert test_duplicate([1,1,2,2,3,3,4,4,5])==True \end{lstlisting} \\ \cmidrule{3-3}

Stage-1 & \colorbox{red!30}{\xmark} &
\begin{lstlisting}[style=python]
def test_duplicate(arraynums):
    assert test_duplicate(arraynums) == False \end{lstlisting} \\ \cmidrule{3-3} 

Stage-2 & \colorbox{red!30}{\xmark} &
\begin{lstlisting}[style=python]
def test_duplicate(arraynums):
    for i in arraynums:
        if i in arraynums:
            return True
    return False \end{lstlisting} \\ \cmidrule{3-3}
 
FT & \colorbox{green!25}{\cmark} &
\begin{lstlisting}[style=python]
def test_duplicate(arraynums):
    return len(arraynums)!= len(set(arraynums)) \end{lstlisting} \\
\bottomrule
\end{tabular}
\caption{Example 1 from MBPP}
\label{tab:example_1_mbpp}
\end{table*}

\begin{table*}[!h]
\lstset{linewidth=9.1cm}
\centering
\begin{tabular}{>{\footnotesize\sc}l>{\footnotesize\sc}cl}
\toprule
Prompt & &
\begin{lstlisting}[style=examples]
Write a function to calculate the harmonic sum of n-1. 
Your code should pass these tests:
assert harmonic_sum(7) == 2.5928571428571425
assert harmonic_sum(4) == 2.083333333333333
assert harmonic_sum(19) == 3.547739657143682 \end{lstlisting} \\ \cmidrule{3-3}

Stage-1 & \colorbox{red!30}{\xmark} &
\begin{lstlisting}[style=python]
def harmonic_sum(n):
    assert n > 0
    assert n <= 1
    assert n % 2 == 0
    assert n > 1
    assert n % 3 == 0
    assert n > 2
    assert n % 5 == 0
    assert n > 3
    assert n % 7 == 0
    assert n > 4
    assert n % 11 == 0
    assert n > 5
    assert n % 13 == 0
    assert n > 6
    assert n % 15 == 0
    assert n > 7
    assert n % 17 == 0
    assert n > 8
    assert n % 19 == 0
    assert n > 9
    assert n % 23 == 0
    assert n > 10
    assert n % 29 == 0
    assert n > 11
    assert n % 31 == 0
    assert n > 12
    assert n % 41 == 0
    assert n > 13
    assert n % 43 == n
    assert n > 14 \end{lstlisting} \\ \cmidrule{3-3} 

Stage-2 & \colorbox{green!25}{\cmark} &
\begin{lstlisting}[style=python]
def harmonic_sum(n):
    return sum([i for i in range(1, n + 1)]) \end{lstlisting} \\ \cmidrule{3-3}
 
FT & \colorbox{red!30}{\xmark} &
\begin{lstlisting}[style=python]
def harmonic_sum(n):
    return (n * (n + 1)) / 2 \end{lstlisting} \\
\bottomrule
\end{tabular}
\caption{Example 2 from MBPP}
\label{tab:example_2_mbpp}
\end{table*}

\end{document}